%% file: iclr2027_conference.tex
\documentclass{article} 
\usepackage{iclr2027_conference,times}

\input{math_commands.tex}

\usepackage{hyperref}
\usepackage{url}

\usepackage{booktabs}
\usepackage{amsmath} 
\usepackage{tabularx}
\usepackage{graphicx}
\usepackage{adjustbox}
\usepackage{multirow}
\usepackage{caption}
\usepackage{threeparttable}
\usepackage{fontawesome5}

\makeatletter
\newcommand{\appendixcontents}{%
  \section*{Appendix Contents}%
  \setcounter{tocdepth}{2}%
  \@starttoc{atoc}}
\makeatother

\makeatletter
\renewcommand{\@maketitle}{%
  \vbox{\hsize\textwidth
    \centering
    \vskip 0.08in
    {\LARGE\bfseries \@title\par}
    \vskip 0.16in
    {\normalsize \@author\par}
    \vskip 0.20in
  }%
}
\makeatother

\title{Less Fusion, Same Action: Revisiting Redundant Connectivity in Mixture-of-Transformer based Robot Policy Models}

\author{\textbf{Songyao Jin}$^{1,2,\dagger}$,
  \textbf{Kun Zhou}$^{2,*}$,
  \textbf{Xinyue Wang}$^{1,2}$,
  \textbf{Lukun He}$^{1,2}$,
  \textbf{Jinglin Cao}$^{1,2}$,\\
  \textbf{Lingjun Mao}$^{1,2}$,
  \textbf{Wenpeng Xu}$^{1,2}$,
  \textbf{Fan Feng}$^{1}$,
  \textbf{Biwei Huang}$^{1,2}$\\[5pt]
    $^1$ University of California San Diego
    \qquad
    $^2$ Aether AI \\
    {
    \small
    $^\dagger$ Work done during internship in Aether AI \qquad 
    $^*$ Corresponding author\\
    \texttt{soj007@ucsd.edu}, \quad \texttt{franciskunzhou@gmail.com}}
}

\providecommand{\avgdelta}[2]{%
  #1\rlap{\hspace{0.3em}{\scriptsize$(#2)$}}%
}

\iclrfinalcopy 
\begin{document}

\maketitle

\fancyhead{}
\renewcommand{\headrulewidth}{0pt}

\begin{abstract}
Mixture-of-Transformers (MoT) robot policies typically maintain dense expert connectivity across network depth, yet it remains unclear how much of this connectivity is actually necessary for downstream adaptation. We conduct a systematic empirical study of expert connectivity redundancy by sparsifying pretrained MoT policies during fine-tuning across multiple architectures and robot manipulation benchmarks. We find that substantial portions of dense expert connectivity can be removed while largely preserving downstream performance, and that different sparse masks with matched budgets achieve similar results, suggesting that no particular subset of expert connections is uniquely indispensable. Motivated by this flexibility, we introduce a structured pruning pattern that concentrates cross-expert connectivity in the early layers of the network. Using early-half fusion as a representative configuration, this design maintains comparable task success while enabling practical deployment simplifications, including truncation of unused conditioning layers and reduction of historical KV-cache storage in autoregressive policies. Deployment measurements further show reductions in loaded parameters, GPU memory, and per-chunk latency. Together, our results suggest that dense expert connectivity in pretrained MoT robot policies is substantially redundant, and that structured pruning provides a simple path toward more efficient deployment without training a new model from scratch. 
\end{abstract}

\begin{center}
\small
\href{https://github.com/Songyao-Jin/less-fusion-same-action}{%
  \faGithub\ \textbf{Code}
}
\qquad
\href{https://songyao-jin.github.io/less-fusion-same-action/}{%
  \faGlobe\ \textbf{Project Page}
}
\qquad
\href{https://huggingface.co/collections/Songyao-Jin/less-fusion-same-action-6b0e85b186dd80488c05bb77}{%
    \raisebox{-0.2\height}{%
    \includegraphics[height=1.1em]{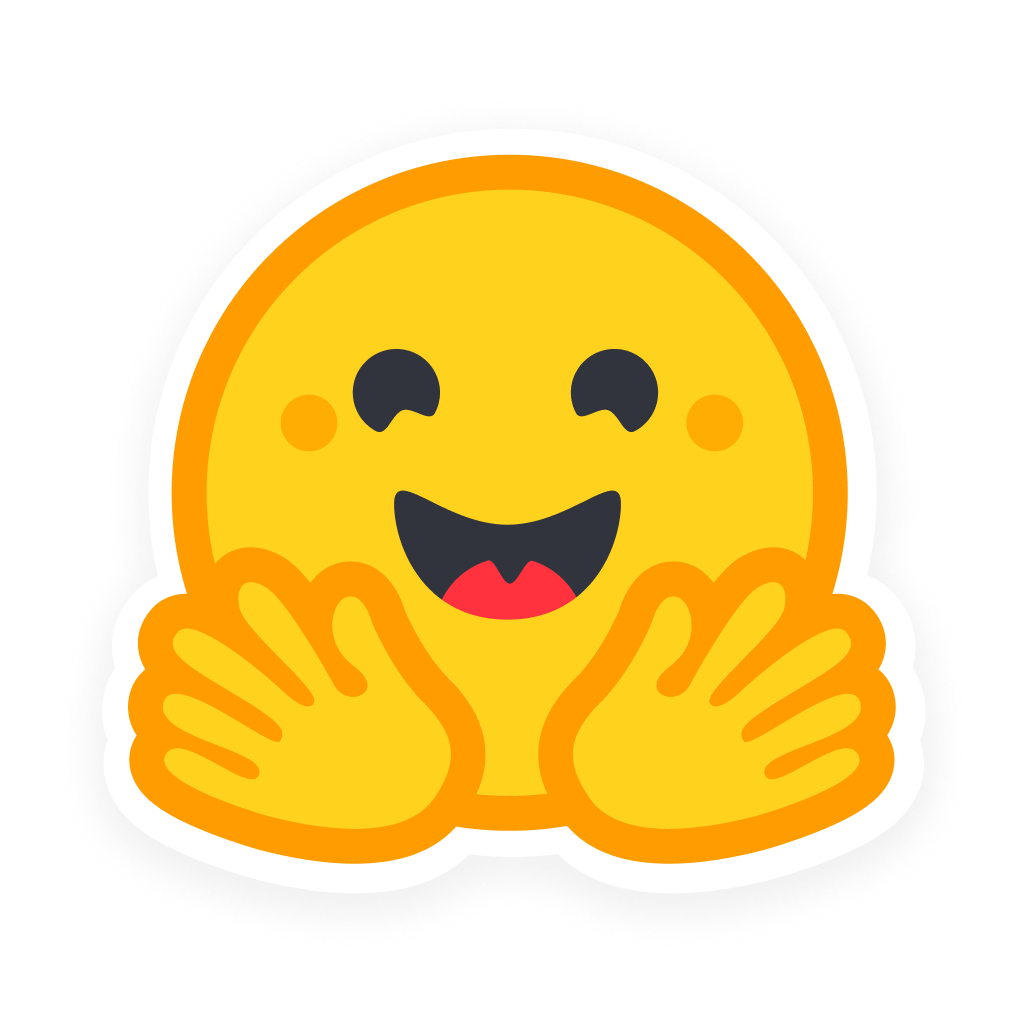}%
  }\ \textbf{Model}
}
\end{center}

\vspace{-0.5em}

\section{Introduction}
Recent advances in vision-language-action models and world-action models have increasingly adopted Mixture-of-Transformers (MoT) architectures \citep{liang2024mixture} to integrate heterogeneous capabilities for robotic control \citep{black2024pi_0,intelligence2025pi_,cai2026internvla,bi2026motus}. In these models, functionally specialized expert streams are responsible for different roles, such as visual-language understanding, future-state prediction, and action generation, while exchanging information through joint attention across Transformer layers. This layer-wise expert fusion provides a flexible interface for combining pretrained representations and task-specific prediction modules, and has become a common design choice in recent robot foundation policies.

Despite its widespread use, the connectivity structure of expert fusion remains poorly understood. Existing MoT policies typically preserve their native expert interactions throughout most or all of the network depth \citep{intelligence2025pi_,nvidia2026cosmos3,luo2026being,li2026causal,li2026wall}, implicitly assuming that dense communication across layers is necessary for effective downstream control. However, it is unclear whether every expert interaction must be retained after pretraining, or whether substantial portions of this connectivity are redundant during downstream adaptation. This raises a fundamental question: \emph{how much expert communication does a pretrained MoT robot policy actually need, and at which layers is such communication necessary?}

To answer this question, we conduct an empirical study on expert connectivity redundancy by systematically sparsifying the native expert-fusion connections of pretrained MoT policies during downstream fine-tuning. Across multiple representative architectures and robot manipulation benchmarks, we find that substantial portions of expert connectivity can be removed while retaining performance close to their dense counterparts. More importantly, different sparse masks with the same connectivity budget achieve similar performance, despite preserving substantially different sets of expert interactions. These results suggest that successful downstream adaptation is not dependent on a particular fixed subset of indispensable connections, revealing substantial redundancy and flexibility in how pretrained experts communicate across network depth.

Building on this observation, we further ask whether the flexibility of expert connectivity can be translated into a structured pruning pattern that directly benefits deployment efficiency. We introduce an early-layer fusion design that preserves cross-expert connectivity only in the early layers while removing it from later layers. Using early-half fusion as a representative configuration, this simple structure maintains comparable downstream task performance, while making the upper-layer computation substantially more separable across experts. As a result, unused upper conditioning layers can be removed to reduce deployed parameter count, and for autoregressive policies, restricting upper-layer attention to the current chunk can further eliminate historical KV-cache requirements. These results show that the redundancy identified in our empirical study can be converted into concrete efficiency gains without redesigning or pretraining a new robot foundation policy.

Our main takeaways are:

$\bullet$ \textbf{Expert connectivity is highly redundant.} Pretrained MoT robot policies can tolerate substantial sparsification during downstream adaptation with little performance loss.

$\bullet$ \textbf{No specific sparse topology is indispensable.} Different masks with similar sparsity budgets achieve comparable performance.

$\bullet$ \textbf{Fusion can be concentrated in early layers.} Restricting cross-expert communication to the first half of the network largely preserves downstream performance.

$\bullet$ \textbf{Structured pruning improves deployment efficiency.} Early-layer fusion enables parameter truncation and KV-cache reduction, lowering memory usage and inference latency.

\begin{figure}[t]
    \centering
    \includegraphics[width=\linewidth]{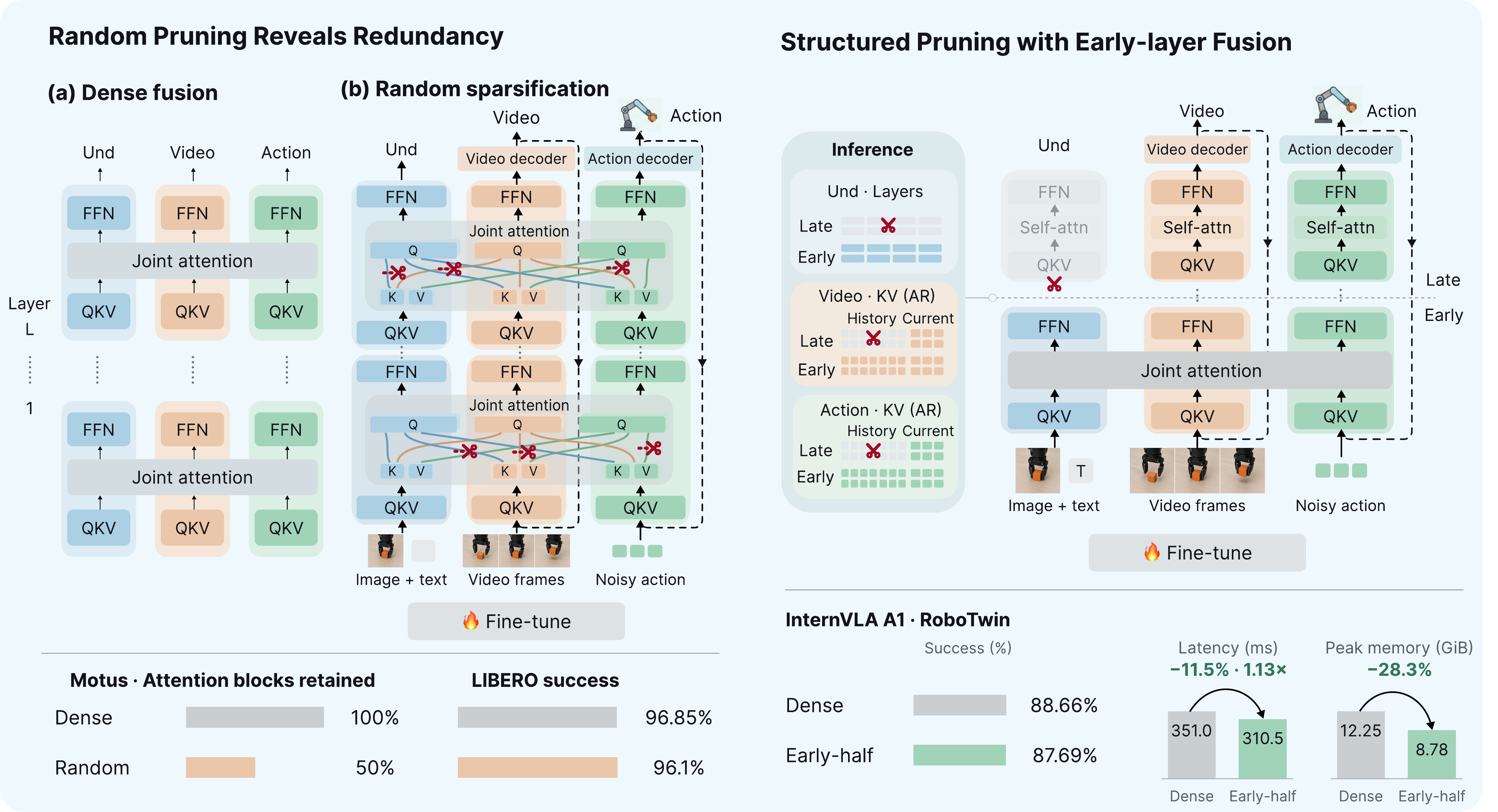}
    \caption{\textbf{Overview of our study.} 
    \textbf{Left:} Random sparsification reveals substantial redundancy in dense expert connectivity while largely preserving downstream performance.
    \textbf{Right:} Early-layer fusion concentrates cross-expert fusion in early layers. Early-half fusion ($N=\lfloor L/2\rfloor$) is used as a representative configuration. While preserving performance, it enables reductions in deployed parameters, KV-cache storage, GPU memory, and inference latency.
    }
    \label{fig:overview}
\end{figure}



\section{Preliminaries}
\label{sec:prelim}

\paragraph{Mixture-of-Transformer based Robot Policy Models.}
The policy models considered in this work primarily follow a MoT design~\citep{liang2024mixture,deng2025emerging}, in which functionally distinct token groups, referred to as \emph{experts}, exchange information through joint attention along each Transformer layer. Let $\mathcal{E}=\{e_1,\dots,e_K\}$ denote the expert set and $L$ the number of Transformer layers, indexed by $\ell\in\{1,\dots,L\}$. The token representations entering layer $\ell$ are organized by expert identity as $\mathbf{X}^{(\ell)}=[\mathbf{X}^{(\ell)}_{e_1};\dots;\mathbf{X}^{(\ell)}_{e_K}]$. In general, the attention projection and feed-forward parameters are expert-specific, although the shared-backbone variants considered below reuse them across streams. Regardless of parameterization, the resulting queries, keys, and values are grouped by their associated experts. For two experts $e_i$ and $e_j$, the attention-logit block at layer $\ell$, through which queries $\mathbf{Q}_{e_i}^{(\ell)}$ read keys $\mathbf{K}_{e_j}^{(\ell)}$, is given by
\begin{equation}
\mathbf{S}_{e_i\rightarrow e_j}^{(\ell)} = \mathbf{Q}_{e_i}^{(\ell)}
\mathbf{K}_{e_j}^{(\ell)\top} 
\big/
\sqrt{d} \quad.
\label{eq2}
\end{equation}
where $d$ is the common query/key dimension of each attention head; the head index is omitted for clarity. The notation $e_i\rightarrow e_j$ indicates that queries from $e_i$ read keys from $e_j$. Diagonal blocks $\mathbf{S}_{e_i\rightarrow e_i}^{(\ell)}$ correspond to within-expert self-attention, whereas off-diagonal blocks $\mathbf{S}_{e_i\rightarrow e_j}^{(\ell)}$, $i\neq j$, enable cross-expert information exchange. 


\begin{table}[t]
\centering
\begin{threeparttable}
\caption{
Architectural variants of the studied robot policies. $^\ast$ denotes the publicly released shared-backbone implementation.
}
\label{tab:model_taxonomy}
\small
\begin{tabular}{lccccc}
\toprule
\textbf{Model}
& \textbf{Type}
& \textbf{Denoising}
& \textbf{Cross-Chunk}
& \textbf{Backbone} 
& \textbf{Expert streams} \\
\midrule
$\pi_{0.5}$
& VLA
& Causal
& Non-AR
& Separate 
& Und, Action\\
InternVLA-A1
& VLA
& Causal
& Non-AR
& Separate 
& Und, Gen, Action
\\
Motus
& WAM
& Joint
& Non-AR
& Separate 
& Und, Video, Action\\
LingBot-VA\tnote{$\ast$}
& WAM
& Causal
& AR
& Shared 
& Video, Action\\
\bottomrule
\end{tabular}


\end{threeparttable}

\end{table}

\paragraph{Expert Connectivity Across MoT Layers.}
At each Transformer layer, a MoT policy defines which expert streams can exchange information through attention. Recall that $\mathbf{S}_{e_i\rightarrow e_j}^{(\ell)}$ denotes the attention-logit block through which queries from expert $e_i$ attend to keys from expert $e_j$ at layer $\ell$. We define the native expert connectivity of a pretrained policy as
\begin{equation}
\Omega=
\{(\ell,i,j)
\mid
\mathbf{S}_{e_i\rightarrow e_j}^{(\ell)}
\text{ is permitted by the native attention structure}.\}
\label{eq:fusion-space}
\end{equation}
Each element $(\ell,i,j)\in\Omega$ therefore represents an expert-level connection at layer $\ell$. Diagonal connections ($i=j$) correspond to within-expert attention, whereas off-diagonal connections ($i\neq j$) enable cross-expert information flow. Depending on the underlying architecture, these connections may be symmetric or directional. For example, a causal attention structure may allow one expert to read from another without permitting the reverse direction. This layer-wise connectivity provides a common abstraction across different MoT architectures. In our study, for each native connection, we associate a binary indicator
$m_{e_i\rightarrow e_j}^{(\ell)}\in \{0,1\}$,
where 1 retains the connection and 0 removes it. The resulting attention block is
\begin{equation} 
\widetilde{\mathbf{S}}_{e_i\rightarrow e_j}^{(\ell)} 
= 
\begin{cases} 
\mathbf{S}_{e_i\rightarrow e_j}^{(\ell)}, & m_{e_i\rightarrow e_j}^{(\ell)}=1, \\ -\infty, & m_{e_i\rightarrow e_j}^{(\ell)}=0. 
\end{cases} 
\label{eq:fusion-mask}
\end{equation} 
The binary mask is applied on top of the native attention structure, so connections disallowed by the original architecture remain unavailable. A complete set of masks therefore specifies which within- and cross-expert interactions are preserved at each layer, providing a unified representation of expert connectivity. As a natural \textit{validity constraint}, each query expert must have at least one accessible key expert at every layer after masking. In particular, if a query expert can attend only to itself at a given layer, its within-expert block must be retained.

\section{Empirical Study on Expert Connectivity Redundancy}
\label{sec:method}
We study whether pretrained MoT robot policies can adapt to sparser expert connectivity while retaining comparable performance. We evaluate this question across multiple representative architectures and manipulation benchmarks using controlled downstream sparsification.

\begin{table*}[t]
\centering
\caption{Fusion sparsification on RoboTwin. Entries are task-averaged success rates on a $0-100$ scale,
with 100 episodes per included task. Easy/Hard denote clean/domain-randomized evaluation. A-C are three random masks per model; 
Bold and underlining mark the best and second-best; parentheses indicate performance retention relative to Dense.
}
\label{tab:fusion-robotwin}

\small
\setlength{\tabcolsep}{5pt}
\renewcommand{\arraystretch}{1.12}

\newcommand{\fusiondelta}[2]{%
  #1\rlap{\hspace{0.3em}{\scriptsize$(#2)$}}%
}

\begin{tabular*}{\textwidth}{@{\extracolsep{\fill}}llcccccc@{\hspace{6pt}}}
\toprule
& & & \multicolumn{4}{c}{\textbf{Random masks}}
& \textbf{Structured} \\
\cmidrule(lr){4-7} \cmidrule(l){8-8}
\textbf{Model}
& \textbf{Setting}
& \textbf{Dense}
& \textbf{A}
& \textbf{B}
& \textbf{C}
& \textbf{Mean}
& \textbf{Early-half fusion} \\
\midrule

$\pi_{0.5}$
& Easy & 80.20 & 78.34 & 78.38 & 79.66 & 78.79 & 77.66 \\
& Hard & 80.50 & 81.22 & 81.28 & 82.38 & 81.63 & 79.54 \\
\addlinespace[2pt]
& \textbf{Avg.}
& \textbf{80.35}
& 79.78 & 79.83 & 81.02
& \fusiondelta{\underline{80.21}}{$99.8\%$}
& \fusiondelta{78.60}{$97.8\%$} \\
\midrule
InternVLA-A1
& Easy & 88.52 & 87.80 & 87.34 & 85.94 & 87.03 & 87.74 \\
& Hard & 88.80 & 87.48 & 87.14 & 86.72 & 87.11 & 87.64 \\
\addlinespace[2pt]
& \textbf{Avg.}
& \textbf{88.66}
& 87.64 & 87.24 & 86.33
& \fusiondelta{87.07}{$98.2\%$}
& \fusiondelta{\underline{87.69}}{$98.9\%$} \\
\midrule

Motus
& Easy & 87.72 & 88.16 & 87.04 & 87.32 & 87.51 & 86.16 \\
& Hard & 85.40 & 87.32 & 87.02 & 85.68 & 86.67 & 84.82 \\
\addlinespace[2pt]
& \textbf{Avg.}
& \underline{86.56}
& 87.74 & 87.03 & 86.50
& \fusiondelta{\textbf{87.09}}{$100.6\%$}
& \fusiondelta{85.49}{$98.8\%$} \\
\midrule

LingBot-VA
& Easy & 91.52 & 86.89 & 87.10 & 86.94 & 86.98 & 92.76 \\
& Hard & 90.86 & 87.22 & 88.30 & 87.72 & 87.75 & 92.34 \\
\addlinespace[2pt]
& \textbf{Avg.}
& \underline{91.19}
& 87.06 & 87.70 & 87.33
& \fusiondelta{87.36}{$95.8\%$}
& \fusiondelta{\textbf{92.55}}{$101.5\%$} \\
\bottomrule
\end{tabular*}
\end{table*}

\subsection{Experimental Setup}
\label{sec:exp-setup}


\paragraph{Models.}
We study four representative pretrained MoT robot policies following the paradigm described in Sec.~\ref{sec:prelim} but with different expert composition, denoising mechanism, temporal dependency, and backbone parameterization: $\pi_{0.5}$~\citep{intelligence2025pi_}, InternVLA-A1~\citep{cai2026internvla}, Motus~\citep{bi2026motus}, and LingBot-VA~\citep{li2026causal}. Their architectural characteristics are summarized in Table~\ref{tab:model_taxonomy}, with detailed descriptions and illustrations provided in Appendix~\ref{app:architectural-variants}.

\paragraph{Benchmarks and Evaluation.}
We evaluate on RoboTwin~\citep{chen2025robotwin} and LIBERO~\citep{liu2023libero}. RoboTwin evaluates bimanual manipulation under both Easy and Hard settings, while LIBERO covers language-conditioned manipulation across the Spatial, Object, Goal, and Long suites. In addition, we test $\pi_{0.5}$ on RoboCasa365~\citep{nasiriany2026robocasa365} and MolmoSpaces~\citep{deshpande2026molmob0t}.
We use task success rate on a 0-100 scale as the primary metric. Within each model and evaluation setting, dense and sparse variants are evaluated on matched task sets under identical conditions, with benchmark-specific rules reported alongside the results.


\paragraph{Sparse Adaptation.}
For each model and downstream setting in the study, we start from a dense checkpoint obtained either from an official release or through our own baseline training. Unless otherwise stated, Dense denotes direct evaluation of this checkpoint. The sparse variants are initialized from the corresponding dense checkpoints and continue fine-tuning under their respective masks, which remain fixed at inference. The training data, objective, and fine-tuning configuration are virtually the same as those of their dense baselines. This setup evaluates whether sparse adaptation can attain performance comparable to the dense reference. Model-specific mask constructions and training configurations are detailed with the corresponding experiments and in Appendices~\ref{app:sparsification-details} and~\ref{app:training-details}. 



\subsection{Random Sparsification Without Performance-Based Search}
\label{sec:exp-random}
We first test whether substantial expert connectivity can be removed without searching for a favorable sparse topology. Starting from each dense checkpoint, we randomly sparsify its native expert connections and fine-tune the resulting policy under the fixed sparse structure.
\begin{equation}
\rho_i
=
\frac{1}{|\Omega_i|}
\sum_{(\ell,j)\in\Omega_i}
m_{e_i\rightarrow e_j}^{(\ell)}.
\label{eq:fusion-retention}
\end{equation}
This quantity measures the fraction of native within- and cross-expert connections retained for query expert $e_i$, with $\Omega_i=\{(\ell,j)\mid(\ell,i,j)\in\Omega\}$. The dense policy has $\rho_i=1$ for all experts. We construct random masks independently for each sparsifiable query stream and retain half of its feasible native connections ($\rho_i=0.5$), with both within- and cross-expert connections participating in sampling. This avoids concentrating sparsification on a particular expert and provides a common query-wise retention budget across architectures. All masks satisfy the native attention constraints defined in Sec.~\ref{sec:prelim}. Query streams that cannot be sparsified under the validity constraint remain unchanged ($\rho_i=1$). For autoregressive LingBot-VA, removing an expert-level connection also removes access to both current-chunk and cached historical keys from that expert. Model-specific constructions are detailed in Appendix~\ref{app:sparsification-details}. We fine-tune and evaluate all four models on RoboTwin and LIBERO, with additional experiments on RoboCasa365 and MolmoSpaces for $\pi_{0.5}$. On RoboTwin, we evaluate three independently sampled masks per model at the same retention ratio to assess sensitivity to the particular sparse topology. For other benchmarks, we use one fixed random mask per model-setting pair. All masks are sampled before fine-tuning without performance-based selection.

\begin{table*}[t]
\centering
\caption{Fusion sparsification on LIBERO. Entries are task-averaged success rates over matched task sets, with 50 episodes per included task. 
The random mask is reused across all four suites. 
}
\label{tab:fusion-libero}

\small
\setlength{\tabcolsep}{5pt}
\renewcommand{\arraystretch}{1.12}

\begin{tabular*}{\textwidth}{@{\extracolsep{\fill}}llccccc@{\hspace{3em}}}
\toprule
\textbf{Model}
& \textbf{Variant}
& \textbf{Spatial}
& \textbf{Object}
& \textbf{Goal}
& \textbf{Long}
& \textbf{Average} \\
\midrule

$\pi_{0.5}$
& Dense
& 98.6 & 99.4 & 97.6 & 91.4 & 96.75 \\
& Random
& 99.2 & 98.8 & 98.4 & 92.6
& \avgdelta{\textbf{97.25}}{$100.5\%$} \\
& Early-half fusion
& 98.6 & 97.4 & 97.4 & 92.8
& \avgdelta{96.55}{$99.8\%$} \\
\midrule

InternVLA-A1
& Dense
& 94.0 & 99.6 & 94.2 & 90.0 & \textbf{94.45} \\
& Random
& 91.4 & 99.6 & 91.8 & 92.6
& \avgdelta{93.85}{$99.4\%$} \\
& Early-half fusion
& 88.0 & 97.6 & 94.0 & 84.8
& \avgdelta{91.10}{$96.5\%$} \\
\midrule

Motus
& Dense
& 98.8 & 99.2 & 94.6 & 94.8 & \textbf{96.85} \\
& Random
& 99.4 & 98.6 & 95.8 & 90.6
& \avgdelta{96.10}{$99.2\%$} \\
& Early-half fusion
& 99.0 & 98.4 & 92.8 & 94.4
& \avgdelta{96.15}{$99.3\%$} \\
\midrule

LingBot-VA
& Dense
& 98.4 & 98.2 & 96.4 & 96.0 & \textbf{97.25} \\
& Random
& 95.8 & 96.8 & 94.0 & 90.6
& \avgdelta{94.30}{$97.0\%$} \\
& Early-half fusion
& 97.6 & 99.2 & 97.6 & 93.6
& \avgdelta{97.00}{$99.7\%$} \\
\bottomrule
\end{tabular*}
\end{table*}


\begin{table*}[t]
\centering
\caption{Fusion sparsification of $\pi_{0.5}$ on RoboCasa365 Atomic-Seen. Success rates are reported for all 18 tasks, with 50 episodes per task. Source-to-destination labels denote pick-and-place tasks.} 
\label{tab:fusion-robocasa365}

\small
\setlength{\tabcolsep}{1.5pt}
\renewcommand{\arraystretch}{1.12}

\newcommand{\rcTaskHeader}[1]{%
  \makebox[0pt][l]{\hspace{-0.4em}\rotatebox{40}{\small\strut #1}}%
}

\begin{tabularx}{\textwidth}{
@{}l@{\hspace{8pt}}
*{18}{>{\centering\arraybackslash}X}
@{\hspace{6pt}}c@{}
}
\toprule
\textbf{Variant}
& \rcTaskHeader{Close blender lid}
& \rcTaskHeader{Close fridge}
& \rcTaskHeader{Close toaster oven door}
& \rcTaskHeader{Coffee mug setup}
& \rcTaskHeader{Navigate kitchen}
& \rcTaskHeader{Open cabinet}
& \rcTaskHeader{Open drawer}
& \rcTaskHeader{Open stand mixer head}
& \rcTaskHeader{Counter to cabinet}
& \rcTaskHeader{Counter to stove}
& \rcTaskHeader{Drawer to counter}
& \rcTaskHeader{Sink to counter}
& \rcTaskHeader{Toaster to counter}
& \rcTaskHeader{Slide dishwasher rack}
& \rcTaskHeader{Turn off stove}
& \rcTaskHeader{Turn on kettle}
& \rcTaskHeader{Turn on microwave}
& \rcTaskHeader{Turn on faucet}
& \textbf{Avg.\ (18)} \\
\midrule

Dense
& 4 & 44 & 14 & 30 & 2
& 40 & 52 & 58
& 66 & 70 & 46 & 84 & 34
& 70
& 18 & 20 & 10 & 52
& \textbf{39.67} \\

Random
& 0 & 64 & 4 & 16 & 0
& 40 & 74 & 24
& 56 & 62 & 48 & 78 & 38
& 72
& 8 & 42 & 10 & 32
& 37.11 \\

Early-half fusion
& 10 & 54 & 8 & 24 & 0
& 50 & 30 & 56
& 66 & 64 & 40 & 78 & 54
& 58
& 22 & 26 & 6 & 40
& 38.11 \\
\bottomrule
\end{tabularx}
\end{table*}


\begin{table*}[t]
\centering
\caption{Fusion sparsification of $\pi_{0.5}$-based policies on five MolmoSpaces evaluation configurations. Entries are success rates, with 100 episodes per configuration. Bold marks the highest average.
}
\label{tab:fusion-molmospaces}

\small
\setlength{\tabcolsep}{1pt}
\renewcommand{\arraystretch}{1.12}

\begin{tabular*}{\textwidth}{@{\extracolsep{\fill}}lcccccc@{}}
\toprule
\textbf{Variant}
& \textbf{Pick-MSProc}
& \textbf{Pick-Classic}
& \textbf{Pick\&Place}
& \textbf{P\&P-NextTo}
& \textbf{P\&P-Color}
& \textbf{Avg.} \\
\midrule

Dense
& 77 & 35 & 40 & 29 & 46 & \textbf{45.40} \\

Random
& 82 & 40 & 36 & 30 & 33 & 44.20 \\

Early-half fusion
& 80 & 34 & 42 & 23 & 47 & 45.20 \\

\bottomrule
\end{tabular*}
\end{table*}

\subsection{Result Analysis}
\label{sec:random-results}

\paragraph{Substantial Expert Connectivity Can Be Removed with Limited Performance Loss.}
On RoboTwin (Table~\ref{tab:fusion-robotwin}), success rates averaged over three masks and both evaluation settings retain $99.8\%$, $98.2\%$, $100.6\%$, and $95.8\%$ of the corresponding dense success rates for $\pi_{0.5}$, InternVLA-A1, Motus, and LingBot-VA, respectively. On LIBERO (Table~\ref{tab:fusion-libero}), the corresponding performance retention rates are $100.5\%$, $99.4\%$, $99.2\%$, and $97.0\%$. Across these eight model–benchmark settings, the absolute changes in success rate range from -3.83 to +0.53. Additional evaluations of $\pi_{0.5}$ show the same trend: random sparsification achieves success rates of $37.11$ on RoboCasa365 (Table~\ref{tab:fusion-robocasa365}) and $44.20$ on MolmoSpaces (Table~\ref{tab:fusion-molmospaces}), close to dense references. These results show that removing substantial expert connectivity can largely preserve the performance across architectures and task settings after adaptation.

\paragraph{No Particular Sparse Connectivity Pattern Is Indispensable.}
The RoboTwin experiments further show that different random masks with the same retention budget achieve similar performance, with a maximum success rate difference of 1.31 and standard deviations of 0.32--0.70 in setting-averaged success rates. Despite retaining different subsets of expert interactions, these masks converge to comparable downstream performance after adaptation. This suggests that effective sparse adaptation does not depend on a particular set of expert connections, revealing substantially distributed redundancy in the dense connectivity structure.

\paragraph{Additional Analysis.}
Direct masking without fine-tuning leads to severe performance degradation, while continued dense fine-tuning yields only limited performance changes, highlighting the role of sparse adaptation (Appendices~\ref{app:direct-masking} and~\ref{app:continued-dense}). An experiment with complementary masks provides further evidence of connectivity redundancy and flexibility (Appendix~\ref{app:additional-ablation}). Further uncertainty analysis, including standard deviations across random masks on RoboTwin and repeated evaluations on LIBERO, is provided in Appendix~\ref{app:evaluation-variability}.

\section{Structured Pruning of Expert Connectivity}
\label{sec:exp-structured}
The previous experiments show that effective sparse adaptation does not depend on a particular expert-connectivity pattern. We therefore ask whether this flexibility can be converted into a structured pruning scheme that preserves downstream performance while simplifying deployment. To this end, we introduce a depth-wise pruning pattern that concentrates cross-expert fusion in the lower layers of the network and removes it from upper layers.

\subsection{Pruning MoT with Early-Half Fusion for Efficiency}
\label{sec:early-half}

\paragraph{Early-Layer Fusion.}
We consider pruning with early-layer fusion, which preserves all native expert connectivity in the first $N$ Transformer layers and restricts the remaining layers to within-expert attention. 
For each native attention block $(\ell,i,j)\in\Omega$, the expert-level mask is
\begin{equation}
m_{e_i\rightarrow e_j}^{(\ell)}
=
\begin{cases}
1, & \ell\le N \text{ or } i=j, \\
0, & \text{otherwise}.
\end{cases}
\label{eq:early-half-fusion}
\end{equation}
This structured mask concentrates cross-expert communication in the lower layers of the network, while the upper layers retain only within-expert computation. In contrast to random sparsification, which removes connections at scattered layers, early-layer fusion eliminates cross-expert dependencies from a contiguous upper-layer region. This contiguous structure is particularly useful because once an expert is no longer consumed by any retained upper-layer computation, its corresponding computation or cached states can be removed from deployment.

In our main experiments, we use \textit{early-half fusion} ($N=\lfloor L/2\rfloor$) as a representative configuration. Similarly, for each model, we initialize from the corresponding dense checkpoint and continue fine-tuning with the early-half mask fixed throughout training and inference. 
Random sparsification and early-half fusion serve different purposes: the former probes connectivity redundancy, whereas the latter exploits this redundancy for efficient deployment, with Dense serving as the primary reference for both. Unlike random sparsification, early-half fusion usually only removes cross-expert blocks. We therefore additionally evaluate cross-only random sparsification with the same cross-expert sparsification budget as early-half fusion (Appendix~\ref{app:cross-only-random}).



\paragraph{Deployment Simplification.}
Early-layer fusion enables two forms of deployment simplification, depending on the underlying MoT architecture. For policies with separate conditioning experts, action-only inference no longer requires the upper layers of a conditioning expert once its representations cease to participate in cross-expert fusion. When these layers do not share parameters with retained components, their weights can therefore be excluded from deployment. This allows us to truncate the Und stack in $\pi_{0.5}$, both the Und and Gen stacks in InternVLA-A1, and the Und expert in Motus, while retaining the full depth of diffusion experts needed for iterative denoising. For autoregressive LingBot-VA, the shared Video-Action backbone must retain all layers for current-chunk computation, so expert-layer truncation is not applicable. Instead, we remove cross-chunk KV access beyond the last fusion layer: upper-layer queries attend only to same-expert tokens from the current chunk, while the first $N$ layers retain the native expert connectivity and temporal access. Historical KV states are therefore required only for the first $N$ layers of the network, reducing persistent historical KV storage by half under early-half fusion. Exact layer configurations, retained components, and model-specific implementations are provided in Appendices~\ref{app:sparsification-details} and~\ref{app:deployment-details}.

\subsection{Result Analysis}
\label{sec:deployment-efficiency}

\paragraph{Early-Half Fusion Preserves Comparable Downstream Performance.}
Across architectures, early-half fusion remains close to the original dense policies. On RoboTwin (Table~\ref{tab:fusion-robotwin}), the setting-averaged success rates retain $97.8\%$, $98.9\%$, $98.8\%$, and $101.5\%$ of the corresponding dense success rates for $\pi_{0.5}$, InternVLA-A1, Motus, and LingBot-VA, respectively. On LIBERO (Table~\ref{tab:fusion-libero}), the corresponding performance retention rates are $99.8\%$, $96.5\%$, $99.3\%$, and $99.7\%$. Across these eight settings, the absolute changes in success rate relative to Dense range from $-3.35$ to $+1.36$. Additional evaluations of $\pi_{0.5}$ show the same trend: early-half fusion achieves success rates of $38.11$ on RoboCasa365 Atomic-Seen (Table~\ref{tab:fusion-robocasa365}) and $45.20$ on MolmoSpaces (Table~\ref{tab:fusion-molmospaces}), compared with dense references of $39.67$ and $45.40$, respectively. These results show that cross-expert connectivity need not be maintained throughout the full network depth; concentrating fusion in the first half can largely preserve downstream performance across the evaluated models and settings.

\paragraph{Structured Pruning Reduces Deployment Cost and Improves Inference Efficiency.}
We next measure the deployment benefits enabled by the structured early-half design. Using matched full-depth and pruned implementations on a single NVIDIA L40S (48 GiB), early-half fusion reduces GPU-resident parameters from 3.62B to 2.63B for $\pi_{0.5}$ and from 3.22B to 2.30B for InternVLA-A1, while achieving per-chunk speedups of $1.30\times$ and $1.13\times$, respectively. For LingBot-VA, pruning upper-layer historical dependencies reduces persistent KV storage from 6.72 to 3.36 GiB and peak allocated GPU memory from 20.63 to 17.26 GiB, while reducing median per-chunk latency from 7344 to 5187 ms ($1.42\times$). Measurements use matched model inputs and generation settings with eager PyTorch. Under compiled execution, the speedups reach $1.50\times$ for $\pi_{0.5}$ and $1.19\times$ for InternVLA-A1. Further details are provided in Appendix~\ref{app:deployment-measurements}.


\paragraph{Additional Analysis.}
We further examine alternative fusion patterns under matched sparsification budgets, including alternating, middle-half, and late-half fusion on InternVLA-A1, as well as cross-only random sparsification on selected models. These experiments show that different fusion patterns can support comparable downstream performance, further supporting the observed connectivity redundancy and flexibility. Early-half fusion provides a structured realization of this flexibility that enables practical deployment simplifications. Detailed results are provided in Appendix~\ref{app:fusion-pattern-analysis}.

\begin{table*}[t]
\centering
\caption{
Deployment efficiency of early-half fusion. Params denotes parameters loaded for inference. Prefill measures conditioning computation and is included in total latency; Total reports the median time per action chunk. Speedup is computed from total latency relative to Dense. Peak memory denotes peak allocated GPU memory. Historical KV denotes persistent cross-chunk cache storage at a matched history length. Dash (--) indicates not applicable.
}
\label{tab:fusion-efficiency}

\small
\setlength{\tabcolsep}{3pt}
\renewcommand{\arraystretch}{1.12}

\begin{tabular*}{\textwidth}{@{\extracolsep{\fill}}llcccccc@{}}
\toprule
\textbf{Model}
& \textbf{Variant}
& \textbf{Params}
& \multicolumn{2}{c}{\textbf{Latency (ms)}}
& \textbf{Speedup}
& \textbf{Peak memory}
& \textbf{Historical KV} \\
\cmidrule(lr){4-5}
& & \textbf{(B)}
& \textbf{Prefill}
& \textbf{Total}
& & \textbf{(GiB)} & \textbf{(GiB)} \\
\midrule

$\pi_{0.5}$
& Dense
& 3.62 & 163.7 & 338.4 & $1.00\times$ & 13.76 & -- \\
& Early-half fusion
& 2.63 & 96.0 & 260.1 & $1.30\times$ & 10.05 & -- \\
\midrule

InternVLA-A1
& Dense
& 3.22 & 90.9 & 351.0 & $1.00\times$ & 12.25 & -- \\
& Early-half fusion
& 2.30 & 63.3 & 310.5 & $1.13\times$ & 8.78 & -- \\
\midrule


LingBot-VA
& Dense
& 5.79 & 627 & 7344 & $1.00\times$ & 20.63 & 6.72 \\
& Early-half fusion
& 5.79 & 557 & 5187 & $1.42\times$ & 17.26 & 3.36 \\
\bottomrule
\end{tabular*}
\end{table*}

\section{Further Analysis}

The preceding sections show that connectivity sparsification can be accommodated through downstream adaptation and can be organized into a deployment-friendly structure. We further examine more aggressive sparsification, policies without embodied pretraining, and real-world manipulation.


\subsection{Fusion Is Redundant but Not Dispensable}
\label{sec:fusion-depth}

\begin{figure*}[t]
\centering

\begin{minipage}[t]{0.47\textwidth}
    \vspace{-10pt}
    \centering

    \includegraphics[
        width=\linewidth,
        height=0.17\textheight,
        keepaspectratio
    ]{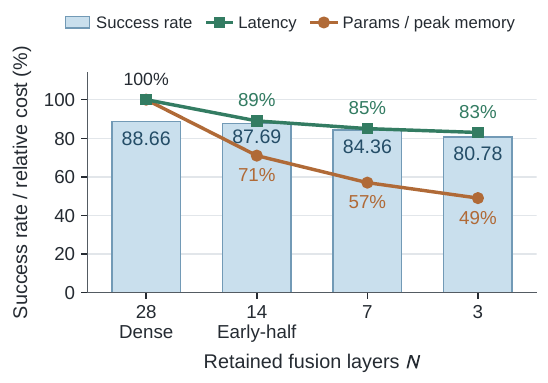}

    \captionsetup{font=small,skip=3pt}
    \captionof{figure}{
    Performance-efficiency trade-off with reduced fusion depth in
    InternVLA-A1. 
    }
    \label{fig:fusion-boundary}
\end{minipage}
\hfill
\begin{minipage}[t]{0.50\textwidth}
    \vspace{0pt}
    \centering

    \captionsetup{font=small,skip=5pt}
    \captionof{table}{
    Fusion sparsification results of Fast-WAM on LIBERO. Entries are task-averaged
    success rates within each task suite, with 50 episodes per included
    task.
    }
    \label{tab:fastwam-libero}

    \small
    \setlength{\tabcolsep}{2.8pt}
    \renewcommand{\arraystretch}{1.16}

    \begin{tabular}{@{}lccccc@{}}
    \toprule
    \textbf{Variant}
    & \textbf{Spatial}
    & \textbf{Object}
    & \textbf{Goal}
    & \textbf{Long}
    & \textbf{Avg.} \\
    \midrule
    Dense
    & 97.8 & 100.0 & 96.4 & 93.0 & 96.8 \\
    Random
    & 96.2 & 97.6 & 94.6 & 89.2 & 94.4 \\
    Early-half fusion
    & 97.4 & 99.8 & 96.8 & 94.4 & 97.1 \\
    \bottomrule
    \end{tabular}

\end{minipage}

\end{figure*}

The strong performance of early-half fusion raises a natural question: how far can expert fusion be reduced before downstream performance begins to degrade? We study this trade-off on InternVLA-A1 by progressively restricting cross-expert communication to the first $N\in\{14,7,3\}$ of its 28 Transformer layers, following the early-layer fusion pattern described in Sec.~\ref{sec:exp-structured}. As shown in Fig.~\ref{fig:fusion-boundary}, retaining fusion in 14 layers reduces the RoboTwin average success rate only from $88.66$ to $87.69$. More aggressive sparsification, however, progressively lowers performance to $84.36$ with 7 fusion layers and $80.78$ with only 3. Meanwhile, deployed parameters decrease to approximately $71\%$, $57\%$, and $49\%$ of Dense, while inference latency decreases more gradually to approximately $89\%$, $85\%$, and $83\%$, respectively. These results reveal a clear performance--efficiency boundary: substantial expert fusion is redundant, but increasingly aggressive sparsification eventually incurs meaningful performance loss despite further reductions in deployed parameters, GPU memory, and inference latency. Details are in Appendix~\ref{app:fusion-budget-analysis}. 



\begin{table}[t]
\centering
\caption{
Real-world task successes out of 20 trials of InternVLA-A1 across five real robot tasks.
}
\label{tab:real-world-success}

\small
\setlength{\tabcolsep}{3pt}
\renewcommand{\arraystretch}{1.16}

\begin{tabular}{@{}lccccc@{}}
\toprule
\textbf{Variant}
& \textbf{Pour Coffee}
& \textbf{Bake Cake}
& \textbf{Pick Bill}
& \textbf{Place Bill}
& \textbf{Change Table Cloth} \\
\midrule
Dense
& 20/20 & 19/20 & 14/20 & 14/20 & 11/20  \\
Random
& 17/20 & 15/20 & 13/20 & 11/20 & 10/20 \\
Early-half fusion
& 20/20 & 19/20 & 12/20 & 13/20 & 9/20 \\
\bottomrule
\end{tabular}%

\end{table}

\begin{figure}[t]
    \centering
     \caption{\textbf{Case study of Place Bill task.}
    From left to right, the robot first pulls the drawer open, places the bill inside, and then pushes the drawer closed. Red circles indicate the bill location.
    }
    \includegraphics[width=\linewidth]{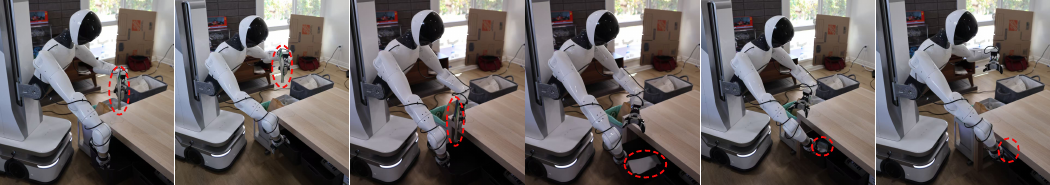}
   
    \label{fig:drawer-bill}
\end{figure}








\subsection{Sparse Fusion Remains Effective Without Embodied Policy Pretraining}
\label{sec:exp-fastwam}

To examine whether sparse expert connectivity relies on embodied-policy pretraining, we additionally evaluate Fast-WAM~\citep{yuan2026fastwamworldactionmodels} on LIBERO. Unlike the pretrained robot policies studied above, Fast-WAM is trained directly on downstream robot data while leveraging pretrained video-model components. We apply the same sparsification framework: random sparsification retains half of the Action-query attention blocks across depth ($\rho_{\mathrm{Action}}=0.5$), with both diagonal and cross-expert blocks participating, while early-half fusion preserves Video-Action fusion in the first 15 of 30 layers and restricts the remaining layers to within-expert attention. Dense, Random, and Early-half variants achieve four-suite average success rates of 96.8, 94.4, and 97.1, respectively (Table~\ref{tab:fastwam-libero}). The comparable performance of the sparse variants suggests that the flexibility of expert connectivity is not limited to pretrained robot policies. Further details are provided in Appendix~\ref{app:fast-wam}.

\subsection{Real-World Experiments}
\label{sec:real-world}
To test whether sparse expert fusion transfers beyond simulation, we deploy InternVLA-A1 on an AgiBot G1 dual-arm robot across five living-room manipulation tasks: Pour Coffee, Bake Cake, Pick Bill, Place Bill, and Change Table Cloth. Policy inference runs on a single NVIDIA RTX~5090, and each variant is evaluated for 20 trials per task without human intervention; a trial is counted as successful if the task goal is achieved at the end of the episode (training and evaluation details in Appendix~\ref{app:real-world}). As shown in Table~\ref{tab:real-world-success}, Early-half fusion matches the dense policy on Pour Coffee and Bake Cake and loses only one to two trials on each of the remaining three tasks, while Random remains within a few trials of Dense on every task. 
These results indicate that the structured early-half pattern preserves most of the dense performance on real hardware.


\section{Related Work}

\paragraph{Mixture-of-Transformers Architecture.}
MoT couples modality- or function-specific Transformer streams through shared attention~\citep{liang2024mixture}, a design closely related to MMDiT~\citep{esser2024sd3}.
LMFusion and BAGEL combine specialized streams for multimodal understanding and generation~\citep{shi2025lmfusion,deng2025emerging}.
Applications of this broader architectural family span text-image-audio~\citep{li2025omniflow}, video-sound~\citep{shan2025hunyuanfoley}, video generation~\citep{bian2025vap}, and omnimodal world modeling~\citep{nvidia2026cosmos3}.
Further applications include 3D generation~\citep{team2025sam3d,lin2026mix3r,yang2026eva01}, human-motion modeling~\citep{li2026llamo}, and time-series forecasting~\citep{wang2026timebraidunifyingtimeseries}.
HBridge further pretrains cross-expert model through intermediate-layer bridges~\citep{wang2026hbridge}.
Recent studies~\citep{zhang2025cross,kaduri2025s,song2026does} report differing findings on the layer-wise importance of cross-modal interactions, with no clear consensus. In contrast, we examine expert connectivity redundancy in pretrained MoT robot policies during downstream adaptation.


\paragraph{Robot Foundation Policies.}
Robot foundation policies connect pretrained representations to actions through token prediction~\citep{brohan2023rt,kim2024openvla} or dedicated diffusion and flow networks~\citep{liu2025rdt,li2024cogact,nvidia2025groot,black2024pi_0,intelligence2025pi_}.
InternVLA-A1 combines understanding, generation, and action experts~\citep{cai2026internvla}, while world-action policies incorporate future visual prediction or video priors~\citep{cen2025worldvla,bi2026motus,li2026causal,ye2026world,kim2026cosmos}.
Recent work also explores intermediate features or learned feature-to-action interfaces~\citep{reuss2025flower,ma2026dit4dit,wang2025vlaadapter}.
Selective fusion also appears in MotuBrain's middle-layer interactions~\citep{team2026motubrain} and SparseMoT's strided video reads with cross-layer KV aggregation~\citep{zhao2026fasterwam_sparsemot}. Unlike these architecture-specific fusion designs, we systematically examine the necessity of native expert connectivity across existing pretrained robot policies.


\paragraph{Efficient Inference and Structural Sparsification of Robot Policies.}
Robot-policy efficiency has been improved through compact backbones~\citep{wen2024tinyvla,shukor2025smolvla}, quantization~\citep{wang2025bitvla,xu2026qvla}, token reduction~\citep{jiang2025lightvla,li2025cogvla,wang2025specprune}, temporal reuse~\citep{xu2025vlacache,tan2025flashvla,li2025spvla}, and faster action decoding~\citep{pertsch2025fast,kim2025oft,song2025pdvla}.
Structural approaches use dynamic layer execution~\citep{yue2024deervla,zhang2025molevla,yang2026dyslvla}, pruning or distillation~\citep{jeon2026shallowpi,jabbour2025gluestick,yang2025efficientvla}, and post-pruning fine-tuning~\citep{chen2025rlrc,nguyen2026clp,sun2026dtr}, with some retaining KV projections from removed language blocks~\citep{sun2026dtr}.
WAM-specific methods additionally reduce future-video generation costs or action-module depth~\citep{yuan2026fastwamworldactionmodels,ma2026fasterwam_dot,li2026efficientwam}. These approaches primarily modify model components, token-level computation, or inference procedures. In contrast, we focus on the redundancy of native expert connectivity while preserving the pretrained expert-block architecture during sparse adaptation.


\section{Conclusion}
We studied whether dense expert connectivity must be preserved throughout network depth when adapting pretrained Mixture-of-Transformers robot policy models to downstream tasks. Across multiple representative architectures and robot manipulation benchmarks, our experiments showed that substantial portions of expert connectivity can be removed during fine-tuning while largely preserving task performance. Moreover, different sparse masks with matched budgets achieved similar results, suggesting that successful adaptation does not depend on a particular fixed subset of expert connections. Building on this observation, we introduced a structured pruning pattern that concentrates cross-expert connectivity in the early layers of the network. With early-half fusion as a representative configuration, this design maintained comparable downstream performance while enabling practical deployment simplifications, including truncation of unused conditioning layers and reduction of historical KV-cache storage in autoregressive policies. Deployment measurements further demonstrated reductions in loaded parameters, GPU memory, and inference latency. 

\subsection*{AI use statement}

We used generative AI tools, including Claude, ChatGPT, and Codex, for language
editing, manuscript organization, LaTeX formatting, literature navigation, and
code generation and debugging. All AI-assisted code was manually reviewed and
tested by the authors before use in experiments. All AI-assisted text, technical
claims, citations, and references in the manuscript were also reviewed and,
where applicable, verified against the original sources and experimental
results. Generative AI was not used as a substitute for experimental validation
or scientific judgment. The authors take full responsibility for the research
methodology, experimental results, analysis, conclusions, and final manuscript.

\subsection*{Ethics statement}

This work studies the efficiency and expert-connectivity structure of existing robot foundation policies. Our experiments do not involve human subjects or the collection of personal or sensitive data. Evaluations are conducted primarily on established simulation benchmarks, with additional real-world robot manipulation experiments performed in a controlled research environment. Real-world experiments are limited to standard manipulation tasks and are conducted with appropriate supervision and hardware safety precautions.

\subsection*{Reproducibility statement}

We provide the model architectures, sparsification formulation, training and evaluation settings, and benchmark protocols needed to reproduce the main experiments in the paper and appendix. Model-specific random and structured mask constructions are described in Appendix~\ref{app:sparsification-details}, and training and evaluation configurations are provided in Appendices~\ref{app:training-details}, \ref{app:fast-wam}, and~\ref{app:real-world}. Additional ablations supporting the analysis of fusion depth are also reported. We use publicly available robot-policy checkpoints and established benchmarks whenever applicable, and report the corresponding evaluation settings and episode counts with each experiment. We plan to release our implementation, sparsification configurations, and supporting experimental data to facilitate reproduction of the reported results.



\bibliography{iclr2027_conference}
\bibliographystyle{iclr2027_conference}

\clearpage   
\appendix

\makeatletter
\let\orig@addcontentsline\addcontentsline
\renewcommand{\addcontentsline}[3]{%
  \def\@tempa{#1}\def\@tempb{toc}%
  \ifx\@tempa\@tempb \orig@addcontentsline{atoc}{#2}{#3}%
  \else \orig@addcontentsline{#1}{#2}{#3}\fi}
\makeatother
\appendixcontents

\clearpage

\section{Architectural Variants}
\label{app:architectural-variants}

\begin{figure}[htpb]
    \centering
    \includegraphics[width=\linewidth]{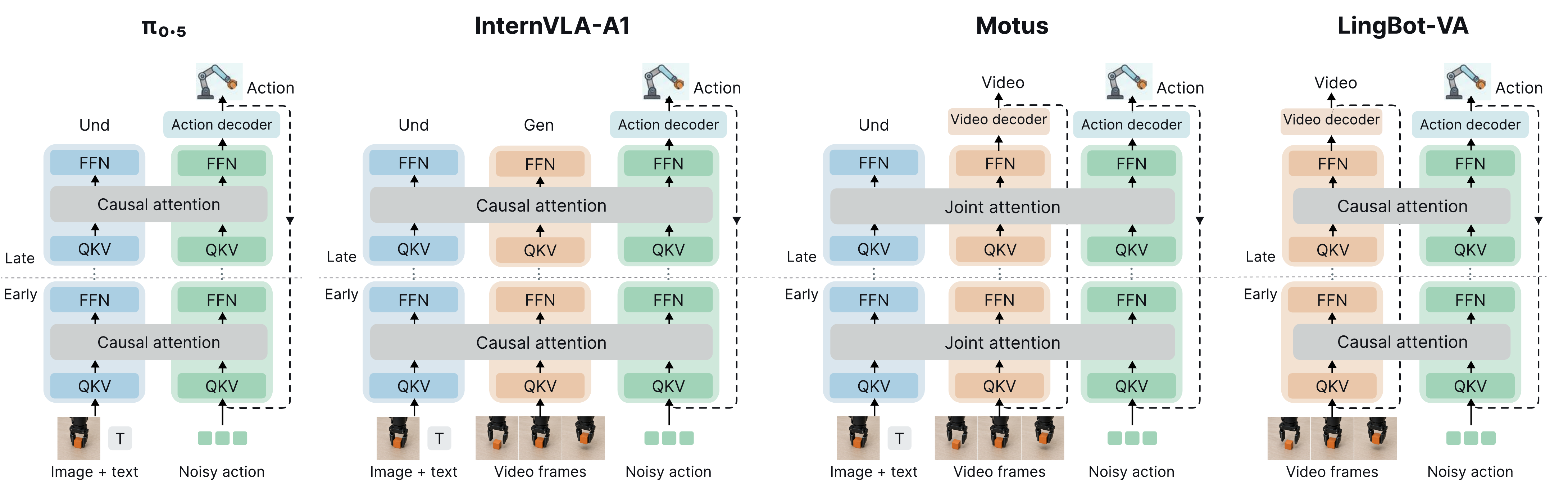}
    \caption{
    \textbf{Architectural overview of the studied MoT robot policies.} The four architectures differ in expert composition, attention connectivity, denoising mechanisms, and backbone parameterization. Dashed arrows denote iterative denoising. LingBot-VA shares backbone parameters across its Video and Action streams, with cross-chunk historical KV
    dependencies omitted for clarity.}
    \label{fig:architectural_overview}
\end{figure}

Recent robot foundation policies instantiate the MoT design by combining an action diffusion expert with other pretrained experts for action generation, while usually differing along several architectural axes. First, \textit{vision-language-action (VLA) models}~\citep{black2024pi_0,intelligence2025pi_,cai2026internvla,luo2026being} emphasize pretrained vision-language representations for semantic understanding, whereas \textit{world-action models (WAM)}~\citep{bi2026motus,li2026causal,ye2026world,li2026wall,team2026motubrain} explicitly incorporate world models for future-state video prediction.
Second, \textit{joint denoising}~\citep{bi2026motus,ye2026world,li2026wall,team2026motubrain} simultaneously denoises multiple diffusion experts, whereas \textit{causal denoising}~\citep{
intelligence2025pi_,cai2026internvla, li2026causal, luo2026being} conditions action denoising on upstream representations computed beforehand, with directed information flow enforced by the native attention mask.
Third, \textit{autoregressive (AR) denoising}~\citep{li2026causal,ye2026world,team2026motubrain} conditions the current chunk on cached keys and values from preceding chunks, whereas \textit{non-autoregressive (Non-AR) denoising}~\citep{black2024pi_0,intelligence2025pi_,cai2026internvla,bi2026motus,luo2026being} operates without such cross-chunk caching. 
Caching the current observation prefix across denoising steps within a single chunk, as in $\pi_{0.5}$, does not constitute cross-chunk autoregression.
We additionally consider shared-backbone variants, in which a single Transformer backbone processes multiple expert token groups~\citep{li2026causal,ye2026world}. These architectural choices are partially coupled. We primarily study four representative policies, $\pi_{0.5}$~\citep{intelligence2025pi_}, InternVLA-A1~\citep{cai2026internvla}, Motus~\citep{bi2026motus}, and LingBot-VA
~\citep{li2026causal},
which span distinct combinations of these choices. Their architectural characteristics are summarized in Table~\ref{tab:model_taxonomy}, and their expert structures are illustrated in Fig.~\ref{fig:architectural_overview}.

\section{Sparsification Details}
\label{app:sparsification-details}

This section describes how the random sparsification and early-half fusion masks in Sec.~\ref{sec:exp-random} and Sec.~\ref{sec:exp-structured} are instantiated for each model. Sparsification only removes native attention paths and never introduces connections prohibited by the original architecture. All masks remain fixed throughout fine-tuning and inference and satisfy the validity constraint defined in Sec.~\ref{sec:prelim}. For random sparsification, the retention budget is enforced across network depth for each sparsifiable query stream. Fig.~\ref{fig:layercut-overview} illustrates early-half fusion and the resulting deployment simplifications across the four studied architectures. The following descriptions refer to the masked architectures before deployment-specific truncation.

\subsection{$\pi_{0.5}$}
\label{app:sparsification-pi05}

$\pi_{0.5}$ combines a vision-language understanding stream (Und) with a flow-matching Action expert. Its native attention mask permits Und queries to attend only to Und keys, while Action queries can attend to both Und and Action keys.

\paragraph{Random Sparsification.}
We leave Und unchanged ($\rho_{\mathrm{Und}}=1$), since removing its only accessible key-expert block would violate the validity constraint. Under the retention budget ($\rho_{\mathrm{Action}}=0.5$) and validity constraint, the Action query stream retains exactly one randomly selected key-expert block (Und or Action) at each layer. Both the cross-expert Und-key block and the within-expert Action-key block participate in sampling. Native attention constraints within the retained blocks remain unchanged.

\paragraph{Early-half Fusion.}
We retain the native connectivity in the first half of layers. In the second half, Action queries attend only to Action keys, while Und queries continue to attend only to Und keys. Thus, Action-to-Und attention is restricted to the first half, and within-expert attention remains available throughout.

\subsection{InternVLA-A1}
\label{app:sparsification-internvla}

InternVLA-A1 contains understanding (Und), generation (Gen), and Action streams. Und and Gen use non-diffusion vision-language components, whereas Action is a diffusion expert. Its native connectivity is causal: Und queries attend only to Und keys, Gen queries can attend to Und and Gen keys, and Action queries can attend to Und, Gen, and Action keys.

\paragraph{Random Sparsification.}
We leave Und unchanged ($\rho_{\mathrm{Und}}=1$), since removing its only accessible key-expert block would violate the validity constraint. We construct separate masks for the Gen and Action query streams, retaining exactly half of each stream's native attention blocks across depth ($\rho_{\mathrm{Gen}}=\rho_{\mathrm{Action}}=0.5$). Both diagonal within-expert and off-diagonal cross-expert blocks participate in sampling, subject to the validity constraint. Native causal masking remains unchanged.

\paragraph{Early-half Fusion.}
We preserve all native connections in the first half of layers. In the second half, Gen queries attend only to Gen keys and Action queries attend only to Action keys; Und remains unchanged throughout. Consequently, cross-expert attention is restricted to the first half, while all within-expert blocks are retained.

\begin{figure}[htpb]
    \centering
    \includegraphics[width=\linewidth]{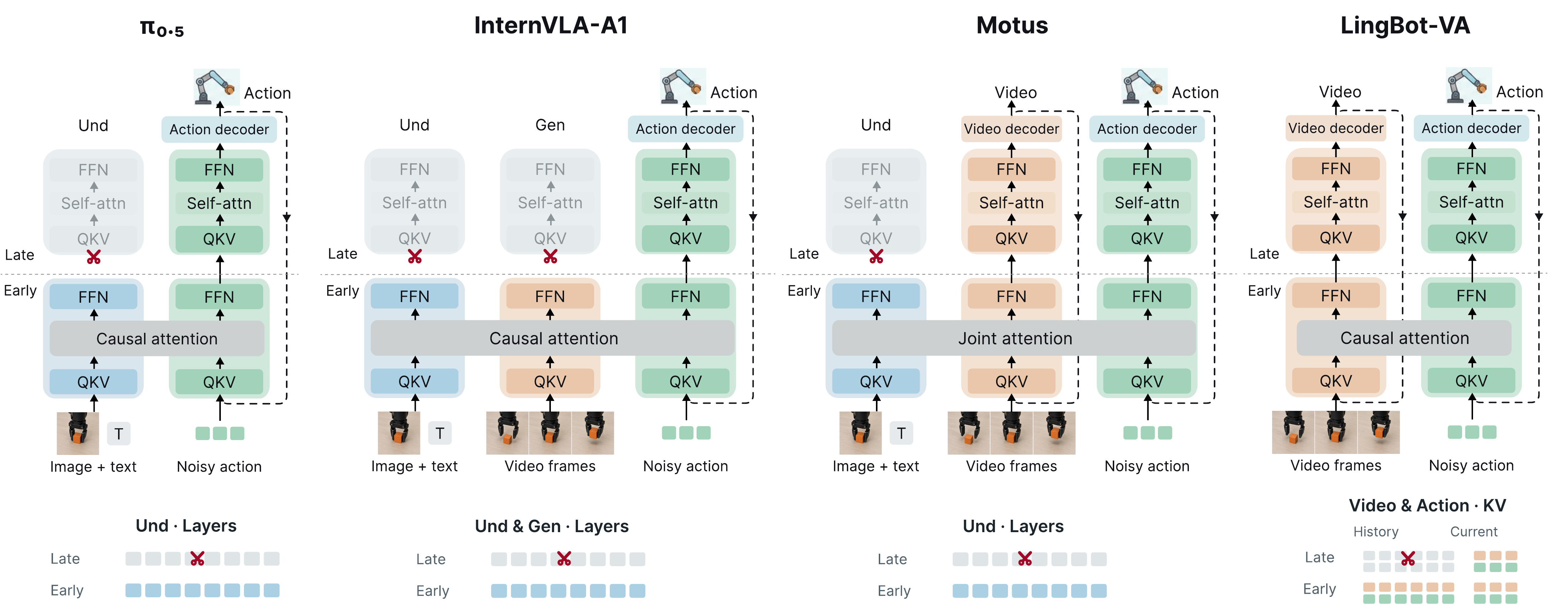}
    \caption{
    \textbf{Structured pruning with early-half fusion across MoT robot policies.}
    Early-half fusion removes cross-expert connectivity in the upper half, enabling conditioning-layer truncation in $\pi_{0.5}$, InternVLA-A1, and Motus. For LingBot-VA, all backbone layers are retained, while restricting upper-layer historical attention enables historical KV-cache reduction. Faded blocks and red scissors indicate components or cached states removed during deployment.
    }
    \label{fig:layercut-overview}
\end{figure}

\subsection{Motus}
\label{app:sparsification-motus}

Motus contains an Und expert operating on upstream VLM features and two diffusion experts, Video and Action. Video and Action are jointly denoised, while all three streams participate in joint attention: each query expert can attend to Und, Video, and Action keys.

\paragraph{Random Sparsification.}
For each query expert, we construct a separate mask over its layer-indexed Und-, Video-, and Action-key blocks. Both diagonal within-expert and off-diagonal cross-expert blocks participate in sampling. Each query stream retains exactly half of its native attention blocks across depth ($\rho_i=0.5$ for all three experts), subject to the validity constraint.

\paragraph{Early-half Fusion.}
We retain full three-expert connectivity in the first half of layers. In the second half, all off-diagonal blocks are removed, leaving only Und-to-Und, Video-to-Video, and Action-to-Action attention. These upper layers therefore introduce no further cross-expert attention.

\subsection{LingBot-VA}
\label{app:sparsification-lingbot}

LingBot-VA contains Video and Action streams that share a diffusion backbone. Video is generated before Action, and both streams may attend to KV states from preceding chunks according to the native temporal mask. Current Action queries can additionally attend to the current predicted video.

\paragraph{Random Sparsification.}
We construct separate masks for the Video and Action query streams, with $\rho_{\mathrm{Video}}=\rho_{\mathrm{Action}}=0.5$. Under the retention budget and validity constraint, each query expert retains exactly one randomly selected key-expert category (Video or Action) at each layer. Dropping a key-expert category removes all otherwise accessible keys from that expert for the corresponding query stream at that layer, including historical and current-chunk keys when applicable.

The masking decision depends only on expert identity, not on temporal distance, frame identity, or cache window. Native temporal masking within each retained category remains unchanged. A category excluded for one query stream may still be accessed by the other; its historical KV cache must remain available whenever a retained computation requires it.

\paragraph{Early-half Fusion.}
We preserve native Video--Action connectivity and cross-chunk KV access in the first half of the shared backbone. In the second half, each query attends only to same-expert tokens within the current chunk: Video queries retain current Video keys, and Action queries retain current Action keys.

All upper-half historical KV access is disabled, including access to same-expert history. This additional temporal restriction eliminates the need to maintain a persistent historical cache for those layers. Native attention constraints within the retained current-chunk blocks remain unchanged.

\subsection{Block-Level Sparsification Budgets}
\label{app:sparsification-budgets}

Random sparsification and early-half fusion serve different purposes in our study. The former examines the redundancy of native expert connectivity, while the latter exploits this redundancy for efficient deployment. Both are primarily evaluated against their corresponding dense policies rather than against each other. Table~\ref{tab:sparsification-budgets} summarizes their block-level sparsification budgets across models.

\begin{table}[htbp]
    \centering
    \caption{
    \textbf{Block-level sparsification budgets.}
    Entries denote the number of removed expert-level attention blocks.
    Exp. denotes expected counts.
    The additional temporal sparsification in LingBot-VA is not included.
    }
    \label{tab:sparsification-budgets}
    \begin{tabular}{lcccc}
        \toprule
        & \multicolumn{3}{c}{\textbf{Random}}
        & \textbf{Early-half} \\
        \cmidrule(lr){2-4}
        \cmidrule(lr){5-5}
        \textbf{Model} & \textbf{Cross (Exp.)}
        & \textbf{Within (Exp.)} & \textbf{Total}
        & \textbf{Cross only} \\
        \midrule
        $\pi_{0.5}$   & 9  & 9  & 18  & 9  \\
        InternVLA-A1  & 42 & 28 & 70  & 42 \\
        Motus         & 90 & 45 & 135 & 90 \\
        LingBot-VA    & 30 & 30 & 60  & 30 \\
        \bottomrule
    \end{tabular}
\end{table}

As shown in Table~\ref{tab:sparsification-budgets}, random sparsification removes both within- and cross-expert blocks, whereas early-half fusion preserves all within-expert blocks. Although their total sparsification budgets differ, early-half fusion removes the same number of cross-expert blocks as the expected count under random sparsification. For LingBot-VA, early-half fusion additionally removes upper-half historical KV access of all experts, which is not reflected in the expert-level block counts.

\section{Deployment Details}
\label{app:deployment-details}

\subsection{Structural Pruning at Deployment}
\label{app:structural-pruning}

Following the structured pruning illustrated in Fig.~\ref{fig:layercut-overview}, Table~\ref{tab:early-half-layer-retention} summarizes the layer and cache retention configurations enabled by early-half fusion. 


\paragraph{Conditioning-layer Truncation.}
Conditioning-layer truncation applies to action-only deployment, where the omitted representations and final outputs are no longer consumed by the retained inference procedure. For $\pi_{0.5}$, the Und stack can be truncated after the last fusion layer, while the upstream visual encoder and input-processing components remain unchanged. For InternVLA-A1, both Und and Gen can be truncated under the sparse action-only execution path. Their layer parameters are separate, so the unused upper-layer weights of both stacks can be excluded from the deployed model. For Motus, the truncated Und expert operates on features extracted by an upstream VLM; the upstream VLM itself remains unchanged.


\paragraph{Historical KV-cache Retention.}
LingBot-VA's full shared backbone remains active for current-chunk computation. Its lower half retains native historical KV access, whereas its upper half attends only to same-expert tokens within the current chunk. Persistent historical KV is therefore maintained only for the lower half, while current-chunk keys and values remain available throughout the backbone. At a fixed history length, batch size, and numerical precision, equal-sized per-layer caches imply a $50\%$ reduction in persistent historical KV storage. 



\begin{table}[htbp]
    \centering
    \caption{
    \textbf{Layer and cache retention under early-half fusion.}
    Native depth denotes the number of Transformer layers per expert
    or shared backbone. Arrows indicate original-to-retained layer
    counts for conditioning experts or persistent historical KV storage.
    }
    \label{tab:early-half-layer-retention}
    \small
    \setlength{\tabcolsep}{6pt}
    \renewcommand{\arraystretch}{1.15}
    \begin{tabular}{@{}lcll@{}}
        \toprule
        \textbf{Model} & \textbf{Native depth}
        & \textbf{Early-half simplification}
        & \textbf{Full-depth components} \\
        \midrule
        $\pi_{0.5}$ & 18
        & Und: $18\rightarrow9$
        & Action \\
    
        InternVLA-A1 & 28
        & Und, Gen: $28\rightarrow14$
        & Action \\
    
        Motus & 30
        & Und: $30\rightarrow15$
        & Video, Action \\
    
        LingBot-VA & 30 (shared)
        & Historical KV: $30\rightarrow15$
        & Shared backbone \\
        \bottomrule
    \end{tabular}
\end{table}

\subsection{Deployment Measurement Protocol}
\label{app:deployment-measurements}

\paragraph{Hardware and Execution.}
Measurements are performed on an eight-GPU NVIDIA L40S host using a single 48\,GiB GPU per run, with NVIDIA driver version 580.178. The policy batch size is one; LingBot-VA's classifier-free guidance internally uses a video-denoising batch size of two. We use FP32 for $\pi_{0.5}$ and InternVLA-A1 and BF16 for LingBot-VA. Within each model, dense and early-half variants use matched input shapes, action horizons, and denoising schedules.


\paragraph{Inputs and Latency.}
Timing inputs consist of fixed-seed synthetic images and zero-valued proprioceptive states, providing reproducible workloads with matched tensor shapes. We measure model-side latency for generating a complete action chunk from device-resident inputs, including the visual encoding, conditioning, and denoising computations required by the model. Model loading, one-time compilation, data transfer, and environment interaction are excluded. CUDA synchronization is performed at the timing boundaries. For $\pi_{0.5}$ and InternVLA-A1, we report the median of 30 timed calls after 5 warm-up calls. LingBot-VA is evaluated through repeated autoregressive rollouts under a saturated attention window.




\paragraph{Memory and Parameter Accounting.}
Parameter counts include all GPU-resident model parameters, including frozen components, with shared parameters counted once. CPU-resident T5 encoders in Motus and LingBot-VA are excluded from parameter and GPU-memory counts. Peak allocated GPU memory is measured using \texttt{torch.cuda.max\_memory\_allocated}, excluding the CUDA context. Persistent historical KV storage is reported separately but is already included in peak GPU memory.




\paragraph{Compiled Execution.}
Table~\ref{tab:fusion-efficiency} reports eager execution. Since the released $\pi_{0.5}$ inference code uses \texttt{torch.compile}, we also compile the action-sampling functions of $\pi_{0.5}$ and InternVLA-A1 (\texttt{mode="max-autotune"}, FP32 with \texttt{float32\_matmul\_precision="highest"}), keeping the hardware, inputs, checkpoints, and timing protocol unchanged.

As Table~\ref{tab:compile-latency} shows, compilation reduces per-chunk latency by $1.6$--$2.4\times$ across configurations and increases the speedup of early-half fusion from $1.30\times$ to $1.50\times$ for $\pi_{0.5}$ and from $1.13\times$ to $1.19\times$ for InternVLA-A1. By reducing kernel-launch overhead in the denoising loop, compilation increases the relative cost of conditioning, where early-half fusion truncates half of the conditioning Transformer layers.

Compiled and eager actions agree to within $4.8\times10^{-7}$ for $\pi_{0.5}$ and $1.9\times10^{-6}$ for InternVLA-A1 (with its image-tokenizer output held fixed). Peak memory changes by less than 0.3~GiB, and a second run reproduces the reported measurements within 3\%. Enabling TF32 matrix multiplication further reduces latency ($101.8/77.1$~ms for $\pi_{0.5}$ and $95.5/83.0$~ms for InternVLA-A1), at the cost of $10^{-3}$-level action deviations. We therefore report results with TF32 disabled.

\begin{table}[ht]
\centering
\caption{
\textbf{Per-chunk latency under compiled execution.} Measurements use the L40S host of Table~\ref{tab:fusion-efficiency} (FP32, batch size 1, median of 30 timed calls). Conditioning and denoising are timed separately and therefore
do not sum exactly to Total. All latencies are in ms; speedup is relative to Dense.
}
\label{tab:compile-latency}
\small
    \begin{tabular}{@{}llccccc@{}}
        \toprule
        & & \textbf{Eager} & \multicolumn{4}{c}{\textbf{Compiled}} \\
        \cmidrule(lr){3-3}\cmidrule(lr){4-7}
        \textbf{Model} & \textbf{Variant} & \textbf{Total}
        & \textbf{Conditioning} & \textbf{Denoising}
        & \textbf{Total} & \textbf{Speedup} \\
        \midrule
        $\pi_{0.5}$ & Dense
        & 338.4 & 149.7 & 55.2 & 210.0 & 1.00$\times$ \\
        & Early-half fusion
        & 260.1 & 81.6 & 50.8 & 139.9 & 1.50$\times$ \\
        \midrule
        InternVLA-A1 & Dense
        & 351.0 & 68.7 & 85.0 & 155.5 & 1.00$\times$ \\
        & Early-half fusion
        & 310.5 & 51.2 & 78.6 & 131.2 & 1.19$\times$ \\
        \bottomrule
    \end{tabular}
\end{table}



\section{Training and Evaluation Details}
\label{app:training-details}

\subsection{Training Protocol and Configurations}

Table~\ref{tab:app-train-config} summarizes the training configurations of the main experiments, with optimizer and numerical settings provided in Table~\ref{tab:app-optim}.

\paragraph{Protocol.}
For each model and benchmark, we use a dense checkpoint obtained either from an official release or through our own baseline training. Unless otherwise specified, sparse variants are initialized from the corresponding dense checkpoints with freshly initialized optimizers and learning-rate schedules. Fine-tuning follows the corresponding training data, objectives, and optimization recipes, preserving the original parameter-freezing settings. The masks are applied within attention and remain fixed throughout fine-tuning and evaluation.




\begin{table}[htbp]
    \centering
    \caption{
    \textbf{Training configurations of the main experiments.}
    We use officially released checkpoints (Released) where available; otherwise, we train dense baselines (Ours) using the corresponding optimization settings, with training steps in parentheses. Steps indicate sparse fine-tuning updates up to the reported checkpoint. Training steps follow official recipes where applicable or are determined by performance stabilization otherwise. Random (R) and Early-half fusion (E) use the same steps unless specified. Batch sizes are global, and cosine schedule lengths indicate the learning-rate decay horizon.
    }
    \label{tab:app-train-config}
    \small
    \setlength{\tabcolsep}{5pt}
    \renewcommand{\arraystretch}{1.15}
    \begin{tabularx}{\textwidth}{
        @{}lllrr>{\raggedright\arraybackslash}X@{}
    }
        \toprule
        \textbf{Model} & \textbf{Benchmark} & \textbf{Dense checkpoint}
        & \textbf{Steps} & \textbf{Batch} & \textbf{LR schedule} \\
        \midrule
        \multirow{4}{*}{$\pi_{0.5}$}
        & RoboTwin
        & Ours (200k)
        & 200k & 128
        & $5{\times}10^{-5}$
        \\

        & LIBERO
        & Released
        & 30k & 256
        & $5{\times}10^{-5}$, constant, warm-up 10k \\

        & RoboCasa365
        & Released
        & \shortstack[r]{R: 40k\\E: 65k}
        & 64
        & $2.5{\times}10^{-5}\to2.5{\times}10^{-6}$,
          cosine (100k), warm-up 1k \\

        & MolmoSpaces
        & Released
        & \shortstack[r]{R: 55k\\E: 65k}
        & 256
        & $5{\times}10^{-5}$, constant, warm-up 1k \\

        \midrule

        InternVLA-A1
        & RoboTwin
        & Released
        & 200k & 128
        & $5{\times}10^{-5}\to5{\times}10^{-6}$,
          cosine, warm-up 2k \\

        & LIBERO
        & Ours (40k)
        & 50k & 128
        & As RoboTwin, cosine (100k) \\


        \midrule

        Motus
        & RoboTwin
        & Released
        & 40k & 256
        & $5{\times}10^{-5}$, $\approx$constant, warm-up 200 \\

        & LIBERO
        & Ours (15k)
        & 15k & 256
        & As RoboTwin \\

        \midrule

        LingBot-VA
        & RoboTwin
        & Released
        & 50k & 64
        & $10^{-5}$, constant, warm-up 10 \\

        & LIBERO
        & Ours (4k)
        & 4k & 120
        & As RoboTwin \\

        \bottomrule
    \end{tabularx}
\end{table}

\begin{table}[htbp]
    \centering
    \caption{
    \textbf{Optimizer and numerical settings.}
    All models use AdamW, with settings shared across benchmarks
    unless specified. FP32 master denotes BF16 computation with
    FP32 master weights, while BF16 denotes BF16 parameters.
    }
    \label{tab:app-optim}
    \small
    \setlength{\tabcolsep}{5pt}
    \renewcommand{\arraystretch}{1.15}
    \begin{tabular}{@{}lccccc@{}}
        \toprule
        \textbf{Model} & $\boldsymbol{(\beta_1,\beta_2)}$
        & \textbf{Weight decay} & \textbf{Grad.\ clip}
        & \textbf{Precision} & \textbf{EMA} \\
        \midrule
        $\pi_{0.5}$
        & $(0.9,0.95)$ & $10^{-10}$ & 1.0
        & FP32 master & 0.999 (LIBERO) \\
        InternVLA-A1
        & $(0.9,0.95)$ & 0.01 & 1.0
        & BF16 & -- \\
        Motus
        & Release default & 0.01 & --
        & FP32 master & -- \\
        LingBot-VA
        & $(0.9,0.95)$ & 0.1 & 2.0
        & FP32 master & -- \\
        \bottomrule
    \end{tabular}
\end{table}

\subsection{Model-Specific Details}
\label{app:model-training-details}

\paragraph{$\pi_{0.5}$.}
We use the PyTorch trainer of OpenPI. On RoboTwin, where no released checkpoint exists, we train the dense policy from $\pi_{0.5}$-base on the 27,500-episode RoboTwin~2.0 dataset and initialize sparse variants from its final weights. Actions are executed in chunks of 15 of the 50 predicted steps. On LIBERO, RoboCasa365, and MolmoSpaces, sparse variants are fine-tuned from the released \texttt{pi05\_libero}, RoboCasa365, and DeMiAn mix5 checkpoints, respectively, following the corresponding released recipes.

\paragraph{InternVLA-A1.}
We use the released LeRobot-based trainer. On RoboTwin, sparse variants are fine-tuned from the released \texttt{InternVLA-A1-3B-RoboTwin} checkpoint. On LIBERO, the dense policy is trained from \texttt{InternVLA-A1-3B}, and sparse variants are initialized from the corresponding dense checkpoint. The real-world training and evaluation settings are described in Appendix~\ref{app:real-world}.

\paragraph{Motus.}
We use the released DeepSpeed ZeRO-1 trainer with the upstream VLM frozen. On RoboTwin, sparse variants are fine-tuned from the released Stage-3 \texttt{Motus\_robotwin2} checkpoint. On LIBERO, the dense policy is trained from the Stage-2 Motus release for 15k updates, while sparse variants receive a further 15k updates. 




\paragraph{LingBot-VA.}
We use the released FSDP trainer. On RoboTwin, sparse variants are fine-tuned from the released \texttt{posttrain-robotwin} checkpoint. On LIBERO, dense policies are trained separately for each suite from \texttt{lingbot-va-base}, following the reproduced LIBERO-Long recipe. Our dense LIBERO-Long reproduction matches the released checkpoint with a success rate of 96.0. Sparse variants are initialized from the released LIBERO-Long checkpoint for the Long suite and from our dense checkpoints for the other suites, where no released checkpoints are available.



\section{Additional Analysis of Expert Connectivity Redundancy}
\label{app:connectivity-analysis}

We further examine expert connectivity redundancy through direct masking, continued dense fine-tuning, and complementary connectivity masks. Unless otherwise specified, the experiments follow the training and evaluation protocols described in Appendix~\ref{app:training-details}.

\subsection{Direct Masking Without Fine-Tuning}
\label{app:direct-masking}

To examine the importance of downstream adaptation, we directly apply Random and Early-half fusion masks to dense checkpoints without updating their weights. We evaluate InternVLA-A1 and Motus on both RoboTwin and LIBERO.

\paragraph{Closed-loop Evaluation.}
As shown in Table~\ref{tab:direct-masking}, direct masking leads to severe performance degradation on RoboTwin, with success rates below $8.0$ across all tested configurations. On LIBERO, all directly masked variants achieve zero successes across the four suites. In contrast, the corresponding sparse fine-tuning experiments largely preserve downstream performance (Tables~\ref{tab:fusion-robotwin} and~\ref{tab:fusion-libero}). These results highlight the importance of downstream adaptation in accommodating substantial changes to native expert connectivity.

\begin{table}[htbp]
    \centering
    \caption{
    \textbf{Direct masking without fine-tuning.}
    Entries are task-averaged success rates.
    All directly masked variants achieve zero successes across
    all four LIBERO suites.
    }
    \label{tab:direct-masking}
    \small
    \setlength{\tabcolsep}{7pt}
    \renewcommand{\arraystretch}{1.15}
    \begin{tabular}{@{}llcccc@{}}
        \toprule
        & & \multicolumn{3}{c}{\textbf{RoboTwin}} & \textbf{LIBERO} \\
        \cmidrule(lr){3-5}\cmidrule(lr){6-6}
        \textbf{Model} & \textbf{Mask} & \textbf{Easy}
        & \textbf{Hard} & \textbf{Avg.} & \textbf{Avg.} \\
        \midrule
        InternVLA-A1 & Early-half
        & 7.42 & 7.50 & 7.46 & 0.00 \\
        & Random
        & 1.28 & 0.80 & 1.04 & 0.00 \\
        \midrule
        Motus & Early-half
        & 0.12 & 0.06 & 0.09 & 0.00 \\
        & Random
        & 0.02 & 0.04 & 0.03 & 0.00 \\
        \bottomrule
    \end{tabular}
\end{table}

\paragraph{Open-loop Action Prediction.}
To further examine the effect of direct masking, we compare predicted action chunks against the corresponding demonstration actions using the same dense checkpoints and inference procedures, without updating model weights. We report root mean squared error (RMSE), dynamic time warping (DTW), and cosine similarity. RMSE is computed in the model-specific normalized action space, and DTW reports trajectory cost normalized by alignment length.

As shown in Table~\ref{tab:direct-masking-open-loop}, direct masking substantially increases action prediction errors, with RMSE and DTW reaching approximately $4.7$--$39\times$ and $5.5$--$62\times$ their dense values, respectively. These observations are consistent with the severe degradation in closed-loop task performance.

\begin{table}[htbp]
    \centering
    \caption{
    \textbf{Open-loop action prediction errors without fine-tuning.}
    Random and Early-half masks are applied directly to dense checkpoints.
    Cosine similarity is omitted for Motus on RoboTwin because it is
    uninformative under its min--max action normalization.
    }
    \label{tab:direct-masking-open-loop}
    \small
    \setlength{\tabcolsep}{6pt}
    \renewcommand{\arraystretch}{1.15}
    \begin{tabular}{@{}lllccc@{}}
        \toprule
        \textbf{Model} & \textbf{Benchmark} & \textbf{Variant}
        & \textbf{RMSE} $\downarrow$ & \textbf{DTW} $\downarrow$
        & \textbf{Cosine} $\uparrow$ \\
        \midrule
        \multirow{6}{*}{InternVLA-A1} & RoboTwin & Dense
        & 0.036 & 0.082 & 0.997 \\
        & & Early-half
        & 0.427 & 1.354 & 0.870 \\
        & & Random
        & 1.381 & 5.055 & 0.030 \\
        \cmidrule(lr){2-6}
        & LIBERO & Dense
        & 0.131 & 0.243 & 0.988 \\
        & & Early-half
        & 0.745 & 1.870 & 0.600 \\
        & & Random
        & 1.448 & 3.726 & 0.030 \\
        \midrule
        \multirow{6}{*}{Motus} & RoboTwin & Dense
        & 0.014 & 0.030 & -- \\
        & & Early-half
        & 0.178 & 0.657 & -- \\
        & & Random
        & 0.215 & 0.802 & -- \\
        \cmidrule(lr){2-6}
        & LIBERO & Dense
        & 0.213 & 0.472 & 0.972 \\
        & & Early-half
        & 1.011 & 2.606 & 0.170 \\
        & & Random
        & 1.040 & 2.682 & 0.080 \\
        \bottomrule
    \end{tabular}
\end{table}

\subsection{Continued Dense Fine-Tuning}
\label{app:continued-dense}

To examine the effect of additional fine-tuning, we continue training the dense policies without applying sparsification masks, using the same initialization, training data, optimization settings, and fine-tuning steps as the corresponding sparse variants. We conduct these experiments on InternVLA-A1 and Motus across RoboTwin and LIBERO, and on LingBot-VA on LIBERO.

As shown in Table~\ref{tab:continued-dense}, continued dense fine-tuning yields only limited changes in success rates relative to the original dense references, ranging from $-0.70$ to $+0.98$. Meanwhile, the sparse variants remain close to Continued Dense under matched training protocols. These results suggest that the observed performance of sparse variants is not simply attributable to additional fine-tuning. Detailed results for individual settings and suites are provided in
Tables~\ref{tab:continued-dense-robotwin}
and~\ref{tab:continued-dense-libero}.

\begin{table}[htbp]
    \centering
    \caption{
    \textbf{Continued dense fine-tuning under matched training protocols.}
    Continued Dense retains the native connectivity and follows the
    fine-tuning protocol of the corresponding sparse variants.
    Entries are average success rates.
    }
    \label{tab:continued-dense}
    \small
    \setlength{\tabcolsep}{5pt}
    \renewcommand{\arraystretch}{1.15}
    \begin{tabular}{@{}llrrrr@{}}
        \toprule
        \textbf{Model} & \textbf{Benchmark} & \textbf{Dense}
        & \textbf{Continued Dense} & \textbf{Random}
        & \textbf{Early-half} \\
        \midrule
        InternVLA-A1 & RoboTwin
        & 88.66 & 88.72 & 87.07 & 87.69 \\
        & LIBERO
        & 94.45 & 93.75 & 93.85 & 91.10 \\
        \midrule
        Motus & RoboTwin
        & 86.56 & 87.54 & 87.09 & 85.49 \\
        & LIBERO
        & 96.85 & 97.00 & 96.10 & 96.15 \\
        \midrule
        LingBot-VA & LIBERO
        & 97.25 & 96.90 & 94.30 & 97.00 \\
        \bottomrule
    \end{tabular}
\end{table}

\begin{table}[htbp]
    \centering
    \caption{
    Detailed results of continued dense fine-tuning on RoboTwin.
    }
    \label{tab:continued-dense-robotwin}
    \small
    \begin{tabular}{lrrr}
        \toprule
        \textbf{Model} & \textbf{Easy} & \textbf{Hard} & \textbf{Avg.} \\
        \midrule
        InternVLA-A1 & 88.30 & 89.14 & 88.72 \\
        Motus        & 87.62 & 87.46 & 87.54 \\
        \bottomrule
    \end{tabular}
\end{table}

\begin{table}[htbp]
    \centering
    \caption{
    Detailed results of continued dense fine-tuning on LIBERO.
    }
    \label{tab:continued-dense-libero}
    \small
    \begin{tabular}{lrrrrr}
        \toprule
        \textbf{Model} & \textbf{Spatial} & \textbf{Object}
        & \textbf{Goal} & \textbf{Long} & \textbf{Avg.} \\
        \midrule
        InternVLA-A1 & 93.0 & 99.8 & 93.6 & 88.6 & 93.75 \\
        Motus        & 99.6 & 99.6 & 95.0 & 93.8 & 97.00 \\
        LingBot-VA   & 98.0 & 97.6 & 96.8 & 95.2 & 96.90 \\
        \bottomrule
    \end{tabular}
\end{table}

\subsection{Complementary Connectivity Masks}
\label{app:additional-ablation}

To further examine whether effective sparse adaptation depends on a particular subset of expert connections, we evaluate the complement of random mask B from Table~\ref{tab:fusion-robotwin} on InternVLA-A1 and RoboTwin. The complement of Mask B exchanges retained and removed blocks within the Gen and Action sparsification pools, while leaving the Und stream unchanged. Both masks follow the same initialization and fine-tuning protocol.

As shown in Table~\ref{tab:complementary-mask}, Mask B and its complement achieve average success rates of $87.24$ and $86.56$, respectively, both remaining close to the dense reference of $88.66$. These results provide further evidence that effective sparse adaptation is not confined to a particular subset of native expert connections.

\begin{table}[htbp]
    \centering
    \caption{
    \textbf{Complementary connectivity masks on InternVLA-A1 and RoboTwin.}
    Both sparse variants follow the same initialization and training protocol.
    Entries are task-averaged success rates.
    }
    \label{tab:complementary-mask}
    \small
    \setlength{\tabcolsep}{8pt}
    \renewcommand{\arraystretch}{1.15}
    \begin{tabular}{lrrr}
        \toprule
        \textbf{Variant} & \textbf{Easy} & \textbf{Hard} & \textbf{Avg.} \\
        \midrule
        Dense                & 88.52 & 88.80 & 88.66 \\
        Mask B               & 87.34 & 87.14 & 87.24 \\
        Complement of Mask B & 86.16 & 86.96 & 86.56 \\
        \bottomrule
    \end{tabular}
\end{table}

\section{Additional Analysis of Fusion Patterns}
\label{app:fusion-pattern-analysis}

We further examine the flexibility of expert connectivity through alternative structured fusion patterns and cross-only random sparsification, with cross-expert sparsification budgets matched to early-half fusion.

\subsection{Alternative Structured Fusion Patterns}
\label{app:structured-patterns}

To examine whether effective sparse adaptation requires fusion at particular depths, we evaluate three alternative structured patterns on InternVLA-A1: Alternating, which retains native fusion in every other layer; Middle-half, which retains fusion in the middle half of layers; and Late-half, which retains fusion in the last half of layers. Together with Early-half fusion, all four patterns retain native cross-expert connectivity in 14 of the 28 Transformer layers while preserving all within-expert blocks. The initialization and fine-tuning protocols remain unchanged.

As shown in Tables~\ref{tab:structured-patterns-robotwin} and~\ref{tab:structured-patterns-libero}, different fusion placements achieve performance close to Dense. Early-half achieves the highest average success rate among the structured patterns on RoboTwin, whereas Late-half performs best on LIBERO. Notably, Early-half and Late-half retain complementary sets of cross-expert connections, yet both support effective downstream adaptation. These results provide further evidence of connectivity redundancy and flexibility across network depth.

\begin{table}[htbp]
    \centering
    \caption{
    \textbf{Alternative structured fusion patterns on InternVLA-A1 and RoboTwin.}
    All sparse patterns retain 14 fusion layers and preserve all
    within-expert blocks.
    Entries are task-averaged success rates.
    }
    \label{tab:structured-patterns-robotwin}
    \small
    \setlength{\tabcolsep}{8pt}
    \renewcommand{\arraystretch}{1.15}
    \begin{tabular}{@{}lrrr@{}}
        \toprule
        \textbf{Fusion pattern} & \textbf{Easy} & \textbf{Hard} & \textbf{Avg.} \\
        \midrule
        Dense       & 88.52 & 88.80 & 88.66 \\
        \midrule
        Early-half  & 87.74 & 87.64 & 87.69 \\
        Alternating & 87.28 & 87.48 & 87.38 \\
        Middle-half & 87.08 & 86.96 & 87.02 \\
        Late-half   & 86.06 & 86.22 & 86.14 \\
        \bottomrule
    \end{tabular}
\end{table}

\begin{table}[htbp]
    \centering
    \caption{
    \textbf{Alternative structured fusion patterns on InternVLA-A1 and LIBERO.}
    All sparse patterns retain 14 fusion layers and preserve all
    within-expert blocks.
    Entries are task-averaged success rates.
    }
    \label{tab:structured-patterns-libero}
    \small
    \setlength{\tabcolsep}{7pt}
    \renewcommand{\arraystretch}{1.15}
     \begin{tabular}{@{}lrrrrr@{}}
        \toprule
        \textbf{Fusion pattern} & \textbf{Spatial} & \textbf{Object}
        & \textbf{Goal} & \textbf{Long} & \textbf{Avg.} \\
        \midrule
        Dense       & 94.0 & 99.6 & 94.2 & 90.0 & 94.45 \\
        \midrule
        Early-half  & 88.0 & 97.6 & 94.0 & 84.8 & 91.10 \\
        Alternating & 91.0 & 98.6 & 93.6 & 89.8 & 93.25 \\
        Middle-half & 91.8 & 97.4 & 90.6 & 89.8 & 92.40 \\
        Late-half   & 94.4 & 99.0 & 92.6 & 92.8 & 94.70 \\
        \bottomrule
    \end{tabular}
\end{table}

\subsection{Cross-Only Random Sparsification}
\label{app:cross-only-random}

Unlike the random sparsification studied in Sec.~\ref{sec:exp-random}, which removes both within- and cross-expert blocks, cross-only random sparsification preserves all within-expert blocks and randomly removes half of the native cross-expert blocks, matching the cross-expert sparsification budget of early-half fusion. We evaluate this setting on InternVLA-A1, Motus, and LingBot-VA on LIBERO, following the corresponding sparse fine-tuning protocols. 



As shown in Table~\ref{tab:cross-only-random}, both Cross-only Random and Early-half fusion achieve comparable performance relative to Dense, with differences between the two sparse patterns within $1.15$ across the three models. These results further support the redundancy and flexibility of native expert connectivity under matched cross-expert sparsification budgets.

\begin{table}[htbp]
    \centering
    \caption{
    \textbf{Cross-only random sparsification on LIBERO.}
    Cross-only Random matches the cross-expert sparsification budget
    of Early-half fusion while preserving all within-expert blocks.
    Dense and Early-half results are reproduced from
    Table~\ref{tab:fusion-libero}.
    Entries are task-averaged success rates.
    }
    \label{tab:cross-only-random}
    \small
    \setlength{\tabcolsep}{5pt}
    \renewcommand{\arraystretch}{1.15}
    \begin{tabular}{@{}lrrrrrrr@{}}
        \toprule
        & \multicolumn{5}{c}{\textbf{Cross-only Random}}
        & \multicolumn{2}{c}{\textbf{Reference (Avg.)}} \\
        \cmidrule(lr){2-6}
        \cmidrule(lr){7-8}
        \textbf{Model} & \textbf{Spatial} & \textbf{Object}
        & \textbf{Goal} & \textbf{Long} & \textbf{Avg.}
        & \textbf{Early-half} & \textbf{Dense} \\
        \midrule
        InternVLA-A1
        & 90.4 & 98.4 & 93.0 & 87.2 & 92.25
        & 91.10 & 94.45 \\
        Motus
        & 99.8 & 99.4 & 93.2 & 93.2 & 96.40
        & 96.15 & 96.85 \\
        LingBot-VA
        & 97.8 & 98.4 & 94.8 & 93.4 & 96.10
        & 97.00 & 97.25 \\
        \bottomrule
    \end{tabular}
\end{table}

\section{Uncertainty Analysis}
\label{app:evaluation-variability}

We examine performance variability across independently sampled random masks on RoboTwin and across repeated evaluations on LIBERO.

\subsection{Variability Across Random Masks}

As described in Sec.~\ref{sec:exp-random}, we independently sample three random masks (A--C) for each model on RoboTwin, with each mask fixed throughout its corresponding fine-tuning and evaluation. Table~\ref{tab:random-mask-variability} summarizes the mean and standard deviation of the average success rates across these runs. The standard deviations range from $0.32$ to $0.70$, indicating limited performance variation across the sampled connectivity patterns and their corresponding fine-tuning runs.

\begin{table}[htbp]
    \centering
    \caption{
    \textbf{Performance variability across random sparsification runs on RoboTwin.}
    Mean and sample standard deviation of average success rates
    across three independently sampled masks and their corresponding
    fine-tuning runs.
    }
    \label{tab:random-mask-variability}
    \small
    \setlength{\tabcolsep}{10pt}
    \renewcommand{\arraystretch}{1.15}
    \begin{tabular}{@{}lc@{}}
        \toprule
        \textbf{Model} & \textbf{Mean $\pm$ Std.} \\
        \midrule
        $\pi_{0.5}$  & $80.21 \pm 0.70$ \\
        InternVLA-A1 & $87.07 \pm 0.67$ \\
        Motus        & $87.09 \pm 0.62$ \\
        LingBot-VA   & $87.36 \pm 0.32$ \\
        \bottomrule
    \end{tabular}
\end{table}

\subsection{Variability Across Repeated Evaluations}

To examine evaluation variability, we perform two additional LIBERO evaluations for both Random and Early-half fusion on $\pi_{0.5}$ and Motus, using the same fine-tuned checkpoints and evaluation protocols. Together with the original evaluations reported in Table~\ref{tab:fusion-libero}, this provides three evaluation runs per setting.

As shown in Table~\ref{tab:repeated-evaluations}, the standard deviations range from $0.23$ to $0.45$, indicating limited variation across repeated evaluations in the tested settings.

\begin{table}[htbp]
    \centering
    \caption{
    \textbf{Performance variability across repeated evaluations on LIBERO.}
    Each setting is evaluated three times using the same fine-tuned
    checkpoint. We report average success rates and their sample
    standard deviations.
    }
    \label{tab:repeated-evaluations}
    \small
    \setlength{\tabcolsep}{7pt}
    \renewcommand{\arraystretch}{1.15}
    \begin{tabular}{@{}llrrrc@{}}
        \toprule
        \textbf{Model} & \textbf{Variant} & \textbf{Original} & \textbf{Eval 2} & \textbf{Eval 3}
        & \textbf{Mean $\pm$ Std.} \\
        \midrule
        $\pi_{0.5}$ & Random
        & 97.25 & 97.20 & 96.55 & $97.00 \pm 0.39$ \\
        & Early-half
        & 96.55 & 96.30 & 96.75 & $96.53 \pm 0.23$ \\
        \midrule
        Motus & Random
        & 96.10 & 97.00 & 96.65 & $96.58 \pm 0.45$ \\
        & Early-half
        & 96.15 & 96.55 & 96.75 & $96.48 \pm 0.31$ \\
        \bottomrule
    \end{tabular}
\end{table}

\section{Additional Analysis of Structural Pruning}
\label{app:structural-pruning-analysis}

\subsection{Progressive Reduction of Fusion Depth}
\label{app:fusion-budget-analysis}

This section provides the detailed ablations underlying Sec.~\ref{sec:fusion-depth}. All early-layer fusion variants follow the same structured rule as early-half fusion: native expert connectivity is retained in the first $N$ Transformer layers, while subsequent layers preserve only within-expert attention. We evaluate $N \in \{14,7,3\}$ in addition to the 28-layer dense model. All sparse variants follow the same initialization and fine-tuning protocol.

As shown in Table~\ref{tab:fusion-depth}, performance remains close to Dense when fusion depth is reduced from 28 to 14 layers, but degrades increasingly under more aggressive cuts. Reducing fusion depth from 14 to 7 layers incurs an additional drop of $3.33$, and reducing it further from 7 to 3 layers incurs another drop of $3.58$. This trend reveals the performance cost of increasingly aggressive fusion sparsification.

\begin{table}[htbp]
    \centering
    \caption{
    \textbf{Progressive fusion-depth reduction on InternVLA-A1.} $N$ denotes the number of initial Transformer layers retaining native cross-expert connectivity, with $N=14$ corresponding to early-half fusion. Entries are task-averaged success rates on RoboTwin.
    }
    \label{tab:fusion-depth}
    \small
    \setlength{\tabcolsep}{8pt}
    \renewcommand{\arraystretch}{1.15}
    \begin{tabular}{@{}lrrrr@{}}
        \toprule
        \textbf{Variant} & \textbf{$N$} & \textbf{Easy} & \textbf{Hard} & \textbf{Avg.} \\
        \midrule
        Dense              & 28 & 88.52 & 88.80 & 88.66 \\
        \midrule
        \multirow{3}{*}{Early-layer fusion} & 14 & 87.74 & 87.64 & 87.69 \\
                           & 7  & 84.24 & 84.48 & 84.36 \\
                           & 3  & 81.06 & 80.50 & 80.78 \\
        \bottomrule
    \end{tabular}
\end{table}

\subsection{Action-Expert Depth Reduction}
\label{app:action-depth-reduction}

While our structured sparsification preserves the depth of the action expert, an alternative approach to reducing inference computation is to directly shorten the action expert. Inspired by the DoT architecture of Faster-WAM~\citep{ma2026fasterwam_dot}, we additionally construct a shallow-action variant of InternVLA-A1 by reducing its 28-layer action expert to a single layer through parameter averaging and DoT-style connections. The resulting model follows the same RoboTwin fine-tuning protocol.

As shown in Table~\ref{tab:action-depth-reduction}, the shallow-action variant achieves an average success rate of $78.41$, compared with $88.66$ for Dense. Unlike Faster-WAM, which trains a dedicated DoT architecture, this experiment directly compresses an existing pretrained policy. 


\begin{table}[htbp]
    \centering
    \caption{
    \textbf{Action-expert depth reduction on InternVLA-A1.}
    The DoT-inspired variant reduces the 28-layer action expert
    to a single layer.
    Entries are task-averaged success rates on RoboTwin.
    }
    \label{tab:action-depth-reduction}
    \small
    \setlength{\tabcolsep}{8pt}
    \renewcommand{\arraystretch}{1.15}
    \begin{tabular}{@{}lcccc@{}}
        \toprule
        \textbf{Variant} & \textbf{Action depth} & \textbf{Easy} & \textbf{Hard} & \textbf{Avg.} \\
        \midrule
        Dense        & 28 & 88.52 & 88.80 & 88.66 \\
        DoT-inspired & 1  & 77.44 & 79.38 & 78.41 \\
        \bottomrule
    \end{tabular}
\end{table}

\section{Additional Details on Fast-WAM}
\label{app:fast-wam}

\paragraph{Architecture and Sparsification.}
Fast-WAM consists of Video and Action experts with 30 Transformer layers and uses pretrained video-model components without large-scale embodied-policy pretraining. We apply the same expert connectivity sparsification framework as for the pretrained robot policies.

For random sparsification, the Video stream remains unchanged, while the Action query stream retains half of its native attention blocks across depth ($\rho_{\mathrm{Action}}=0.5$), with both within- and cross-expert blocks participating in sampling. For early-half fusion, native expert connectivity is retained in the first 15 layers, while the remaining layers preserve only within-expert attention. The masks remain fixed throughout fine-tuning and evaluation.

\paragraph{Training and Evaluation.}
Both sparse variants are fine-tuned from the released \texttt{libero\_uncond\_2cam224} checkpoint using the released LIBERO training recipe. We train for 10 epochs (approximately 21.7k updates) with a global batch size of 128 and a cosine learning-rate schedule initialized at $10^{-4}$. We use AdamW with the default momentum coefficients of the released recipe, a weight decay of 0.01, gradient clipping at 1.0, and BF16 mixed precision. The final checkpoints are evaluated on the four LIBERO suites with 50 episodes per task.

\section{Real-World Experiment Details}
\label{app:real-world}

\paragraph{Platform.}
The AgiBot G1 is a dual-arm mobile manipulator. In our experiments, the base, torso, and head are fixed, and the policy controls the two 7-DoF arms and their grippers. Actions are 16-D absolute joint targets (seven joint angles and one gripper command per arm), where the gripper command is continuous rather than a binary open/close signal. Observations consist of three RGB cameras: a head camera ($800\times1280$) and one wrist camera per arm ($480\times848$), captured at 30 Hz, together with the current 16-D joint state. Images are resized to $224\times224$ with padding, and the policy receives two frames per camera: the current frame and the frame 15 steps (0.5 s) earlier.

\paragraph{Data.}
We collect teleoperated demonstrations of the five tasks in a living-room environment, totaling 2,496 episodes (1.09M frames, approximately 10 hours). Table~\ref{tab:real-world-data} lists the language instructions and the number of demonstrations for each task. Bake Cake demonstrations use two instruction variants that differ only in the description of the cake. Actions and states are normalized using per-dimension means and standard deviations computed over the training data. To account for the imbalance in demonstration counts across tasks, we sample each task with probability proportional to the square root of its frame count.

\begin{table}[htbp]
    \centering
    \caption{
    \textbf{Real-world tasks and demonstrations.}
    Language instructions and the number of teleoperated
    demonstration episodes for each task.
    }
    \label{tab:real-world-data}
    \small
    \setlength{\tabcolsep}{5pt}
    \renewcommand{\arraystretch}{1.15}
    \begin{tabularx}{\textwidth}{
        @{}l>{\raggedright\arraybackslash}Xr@{}
    }
        \toprule
        \textbf{Task} & \textbf{Instruction} & \textbf{Episodes} \\
        \midrule
        Pour Coffee
        & Pour the coffee beans from the bag into the coffee machine's bean hopper.
        & 1,398 \\

        Bake Cake
        & Put the cake with white icing into the air-fryer basket and return the basket.
        & 116 \\

        Pick Bill
        & Pick up the bill from the sofa.
        & 223 \\

        Place Bill
        & Put the bill in the coffee table drawer.
        & 168 \\

        Change Table Cloth
        & Spread the clean tablecloth on the coffee table.
        & 591 \\
        \midrule
        Total & & 2,496 \\
        \bottomrule
    \end{tabularx}
\end{table}

\paragraph{Training.}
All variants use a global batch size of 128, AdamW with the optimizer settings of InternVLA-A1 in Table~\ref{tab:app-optim}, a cosine learning-rate schedule from $5{\times}10^{-5}$ to $5{\times}10^{-6}$ over 60k updates with a 2k-step warm-up, and an action chunk size of 50. The dense policy is initialized from the released \texttt{InternVLA-A1-3B} checkpoint and fine-tuned on the collected demonstrations, with the checkpoint at 40k updates used as Dense. The Random and Early-half fusion variants are initialized from this checkpoint and fine-tuned for a further 40k updates under their respective masks, with freshly initialized optimizers and learning-rate schedules.

\paragraph{Deployment and Evaluation.}
The policy runs as a server on a single NVIDIA RTX 5090, and the attention masks of the sparse variants remain unchanged from fine-tuning. The robot queries the policy at 3 Hz and executes the returned 50-step action chunks in a receding-horizon manner, committing to at most 30 steps of a chunk and blending over 10 steps when switching to a newer chunk. All controller settings are identical across variants.

Each variant is evaluated for 20 trials per task. Object placements are varied across trials and are not paired across variants. A trial is counted as successful if the task goal is achieved at the end of the episode, without human intervention during the trial.

\end{document}

%% file: math_commands.tex
\usepackage{amsmath,amsfonts,bm}

\def\eqref#1{equation~\ref{#1}}

\def\1{\bm{1}}

\DeclareMathAlphabet{\mathsfit}{\encodingdefault}{\sfdefault}{m}{sl}
\SetMathAlphabet{\mathsfit}{bold}{\encodingdefault}{\sfdefault}{bx}{n}

\newcommand{\E}{\mathbb{E}}

